\documentclass[conference]{IEEEtran}
\IEEEoverridecommandlockouts
\usepackage{cite}
\usepackage{amsmath,amssymb,amsfonts}
\usepackage{graphicx}
\usepackage{textcomp}
\usepackage{xcolor}
\usepackage{algorithm} 
\usepackage{algpseudocode} 
\usepackage{multirow}
\usepackage{caption}
\usepackage{subcaption}
\usepackage{svg}
\usepackage{url}

\def\BibTeX{{\rm B\kern-.05em{\sc i\kern-.025em b}\kern-.08em
    T\kern-.1667em\lower.7ex\hbox{E}\kern-.125emX}}
\begin{document}

\title{Closer through commonality: Enhancing hypergraph contrastive learning with shared groups
\footnotesize \thanks{* Corresponding author}
}

\author{\IEEEauthorblockN{Daeyoung Roh\textsuperscript{1}, Donghee Han\textsuperscript{1}, Daehee Kim\textsuperscript{1}, Keejun Han\textsuperscript{2}\textsuperscript{*}, Mun Yi\textsuperscript{3}\textsuperscript{*}}
\IEEEauthorblockA{Graduate School of Data Science, KAIST, Republic of Korea\textsuperscript{1}\\
School of Computer Engineering, Hansung University, Republic of Korea\textsuperscript{2}\\
Department of Industrial and Systems Engineering, KAIST, Republic of Korea\textsuperscript{3}\\
\{dybroh, handonghee, dh.kim\}@kaist.ac.kr, keejun.han@hansung.ac.kr, munyi@kaist.ac.kr
}}
\maketitle

\begin{abstract}
    Hypergraphs provide a superior modeling framework for representing complex multidimensional relationships in the context of real-world interactions that often occur in groups, overcoming the limitations of traditional homogeneous graphs. However, there have been few studies on hypergraph-based contrastive learning, and existing graph-based contrastive learning methods have not been able to fully exploit the high-order correlation information in hypergraphs. Here, we propose a Hypergraph Fine-grained contrastive learning (HyFi) method designed to exploit the complex high-dimensional information inherent in hypergraphs. While avoiding traditional graph augmentation methods that corrupt the hypergraph topology, the proposed method provides a simple and efficient learning augmentation function by adding noise to node features. Furthermore, we expands beyond the traditional dichotomous relationship between positive and negative samples in contrastive learning by introducing a new relationship of weak positives. It demonstrates the importance of fine-graining positive samples in contrastive learning. Therefore, HyFi is able to produce high-quality embeddings, and outperforms both supervised and unsupervised baselines in average rank on node classification across 10 datasets. Our approach effectively exploits high-dimensional hypergraph information, shows significant improvement over existing graph-based contrastive learning methods, and is efficient in terms of training speed and GPU memory cost. The source code is available at \url{https://github.com/Noverse0/HyFi.git}.


\end{abstract}


\section{Introduction}
\label{sec:intro}
Hypergraphs surpass simple pairwise connections, capturing complex interactions in the real world\cite{yang2022semi, he2023robust, yang2023group, luo2022dualgraph, xie2022contrastive, zhu2021inhomogeneous, saifuddin2023hygnn, li2023coclep, yin2022dynamic}. Although recent advances in graph-based learning have primarily focused on these simple relationships, they often fail to account for the intricate group dynamics inherent in real-world data. In contrast, hypergraphs can go beyond pairwise connections to capture high-order data correlations. Their application across diverse fields such as social network analysis\cite{yang2019revisiting, sun2022ms, sun2023self}, bioinformatics\cite{saifuddin2023seq, zhang2022multiscale, you2022cross, liu2022generating}, and complex system modeling is increasingly prevalent, as they can effectively represent the depth of relationships within these domains. These applications showcase the advantages of hypergraphs in capturing high-order data correlations from real-world scenarios\cite{li2021hyperbolic, vijaikumar2021hypertenet, xia2021self, gu2020quantum, liu2023self}.


\begin{figure}[t!]
  \centering
  \begin{subfigure}[b]{0.56\columnwidth}
    \includegraphics[width=\linewidth]{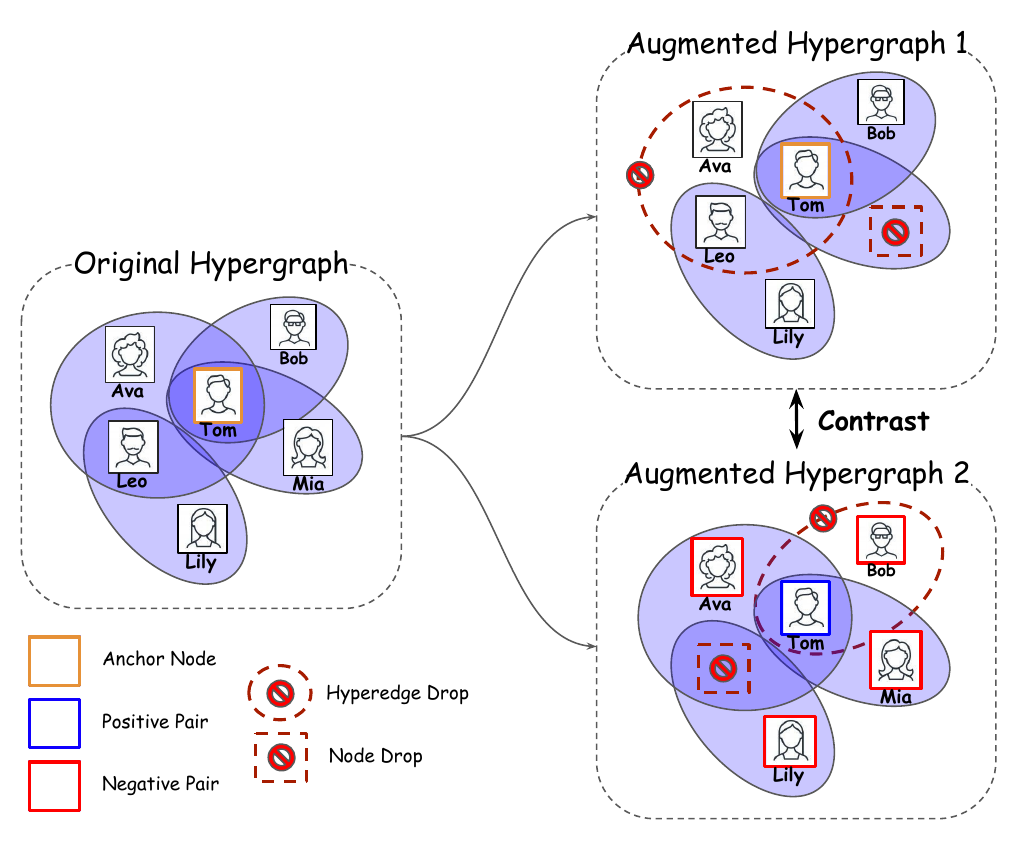}
    \caption{Tranditional HGCL}
  \end{subfigure}
  \begin{subfigure}[b]{0.39\columnwidth}
    \includegraphics[width=\linewidth]{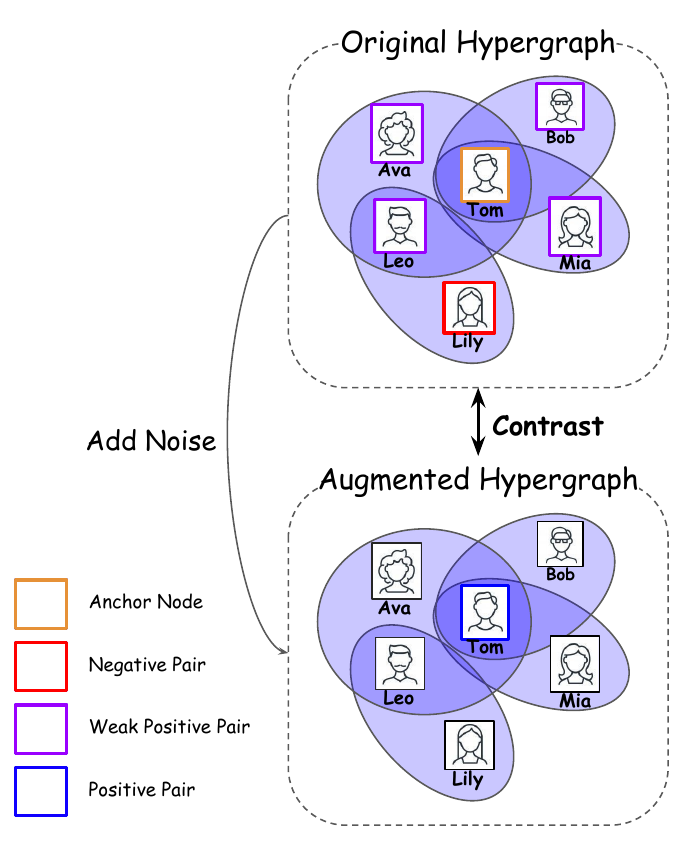}
    \caption{HyFi}
  \end{subfigure}
  \caption{Example of tranditional hypergraph contrastive learning vs. HyFi. In (a), contrastive learning only learns postive or negative pairs between graphs augmented by randomly dropping hyperedges or nodes. In (b), HyFi preserves the topology of the hypergraph by adding noise to the node features, while adding weak postive pair relationships for fine-grained contrastive learning.}
  \label{fig1}
\end{figure}

\begin{figure}[t!]
  \centering
  \begin{subfigure}[b]{0.47\columnwidth}
    \includegraphics[width=\linewidth]{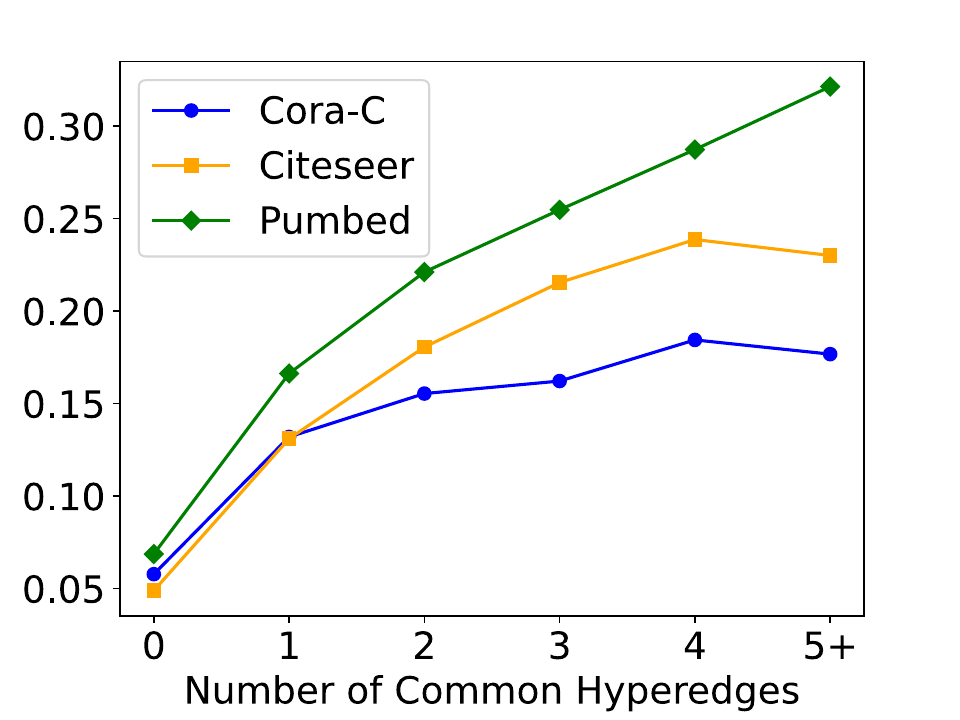}
    \caption{Co-citation}
  \end{subfigure}
  \hspace{5pt} 
  \begin{subfigure}[b]{0.47\columnwidth}
    \includegraphics[width=\linewidth]{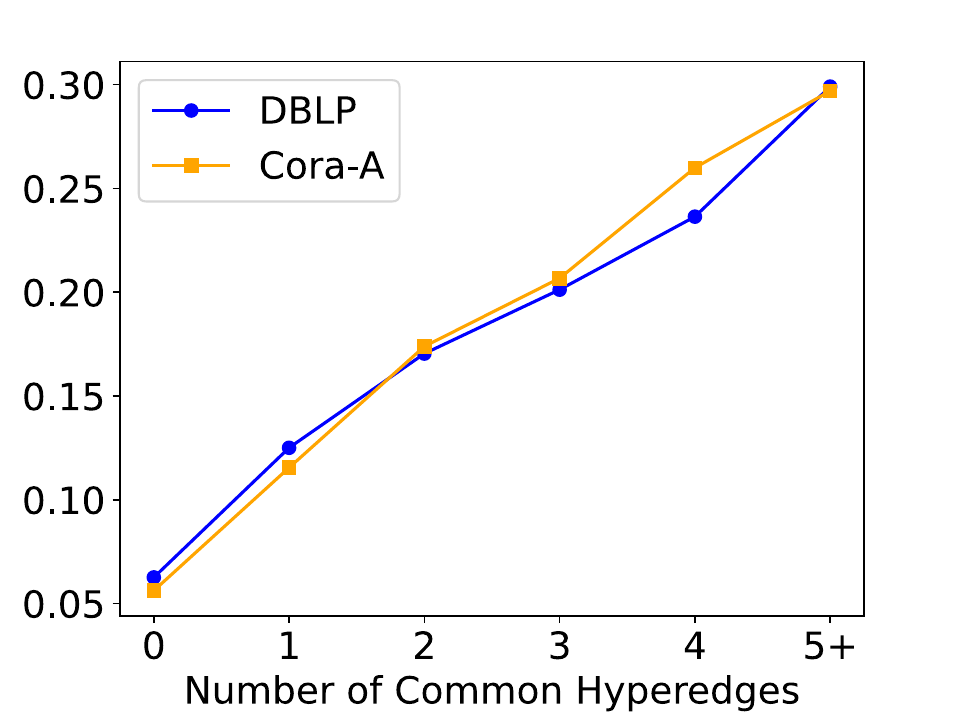}
    \caption{Co-authorship}
  \end{subfigure}
  \begin{subfigure}[b]{0.47\columnwidth}
    \includegraphics[width=\linewidth]{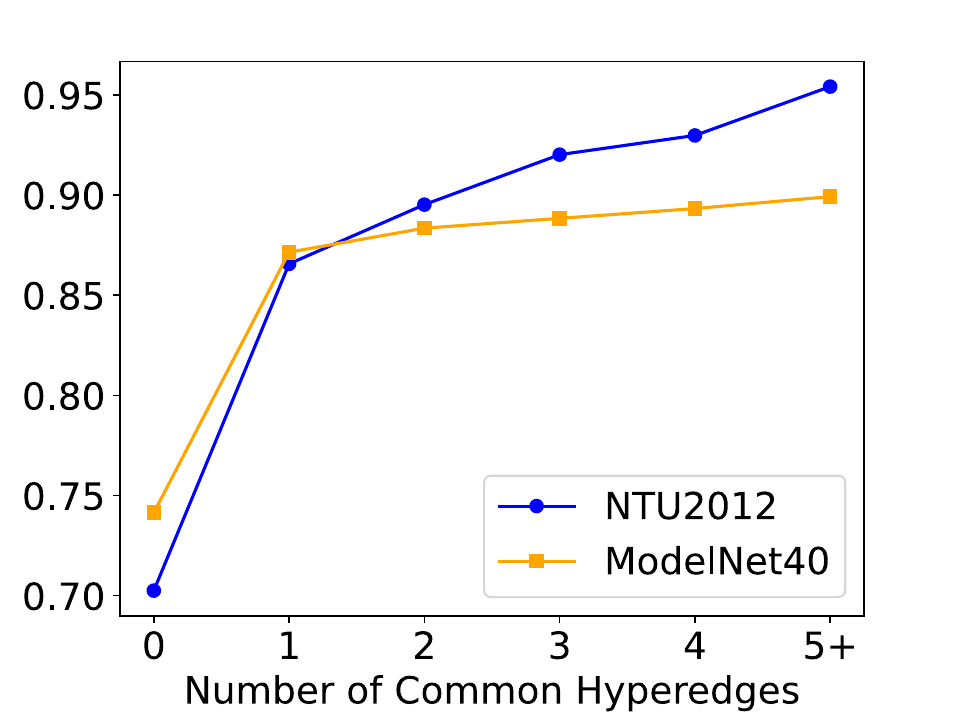}
    \caption{Computer Vision and Graphics}
  \end{subfigure}
  \hspace{5pt} 
  \begin{subfigure}[b]{0.47\columnwidth}
    \includegraphics[width=\linewidth]{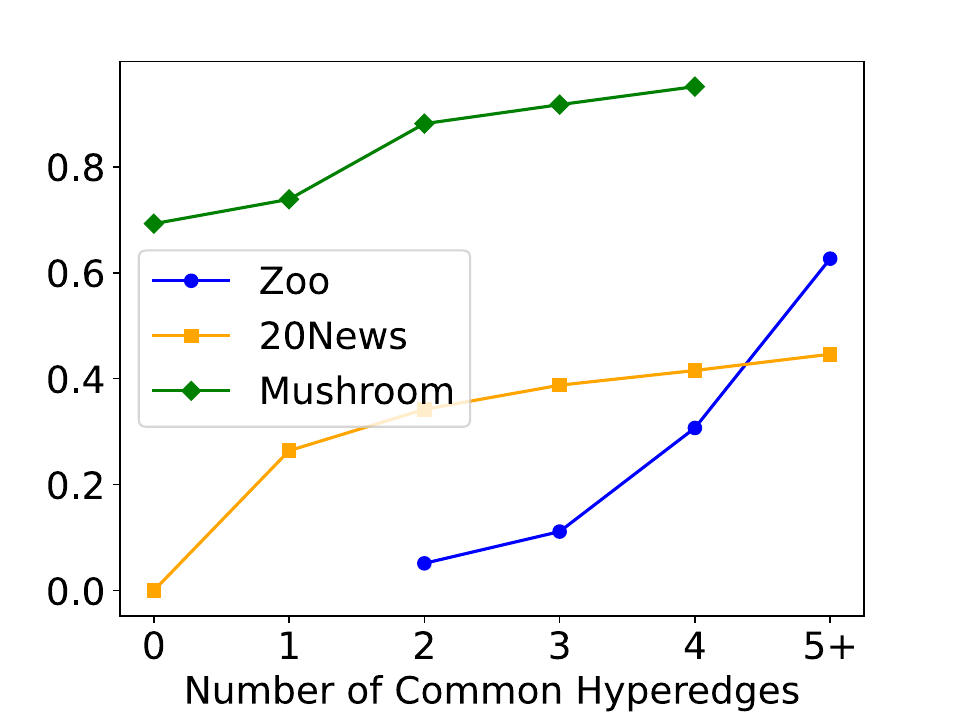}
    \caption{Machine Learning Repository}
  \end{subfigure}
  \caption{The average of the cosine similarity of the node features over the number of commonalities between two nodes, where commonality, defined as a common group, means common hyperedges. The higher the number of common hyperedges, the higher the similarity of the two node features.}
  \label{fig2}
\end{figure}

Despite its potential, research on hypergraph-based learning has been relatively sparse, with even fewer studies exploring contrastive learning, although the possibility of using a hypergraph has recently gained momentum. The complex structure of hypergraphs, which includes nodes and hyperedges, differs significantly from traditional homogeneous graphs, requiring new approaches tailored specifically to hypergraphs for effective learning. Traditional hypergraph contrastive learning (HGCL) relies on contrastive learning between graphs augmented with random variations in nodes or hyperedges. Consequently, it depends heavily on the embedding of the augmented opposite view. In contrast, we propose \textbf{Hy}pergraph \textbf{Fi}ne-Grained contrastive learning (HyFi), a novel method that identifies and contrastively learns new positive pairs within the original view without perturbing the topology of the original graph. Furthermore, our model does not rely on the quality of the augmented view, as it generates positive pairs by simply adding some noise to the node features. Fig. \ref{fig1} illustrates the difference between trainditional HGCL and HyFi.


Traditional graph contrast learning (GCL) typically generates two augmented views of an existing graph and then contrasts them using positive and negative pairs\cite{velivckovic2018deep, peng2020graph, hassani2020contrastive, you2020graph, zhu2020deep}. However, these traditional approaches do not adequately capture the nature of hypergraphs, especially their ability to represent complex group interactions and higher-order correlations represented by commonality. The dichotomous relationship represented by the traditional positive pair and negative pair makes sense for homogeneous graphs with one relation, but cannot capture the higher-order correlations present in hypergraphs with many different relations. Furthermore, efficient contrastive learning is sensitive to the augmentation of the graph, which mostly relies on the randomness of the graph, which can lead to loss of important information in the graph.


Therefore, we propose a new relation, called a weak positive pair, which shows how group sharing can be exploited in contrastive learning. We show that a weak positive pair is a relationship between nodes that share the same group relative to an anchor node, and that we see significant performance gains just by introducing weak positive pairs. Moreover, weak positive pairs can be found within the original view, so there is little dependence on augmenting the graph. These new relationships are based on the idea that we can define ourselves by the people around us who are in the same group as us. In reality, we are strongly influenced by the people around us and relate to similar people.


Furthermore, our approach to augmenting the graph by adding noise to the node features to generate positive samples differs from typical graph augmentation techniques that corrupt the original topology and increase model complexity through random modifications. However, we show that this approach is still sufficient to generate positive samples and is an efficient graph augmentation technique. Therefore, we propose a new form of HGCL method that emphasizes the importance of shared hyperedges and shared nodes while maintaining the structural integrity of the hypergraph. Properly exploiting this commonality is very important in HGCL because hypergraphs are well suited to represent these different relationships through hyperedges.


Distinguishing these types of relationships allows us to capture complex intricacies more accurately. This approach is based on the assumption that the more two nodes share a common group, the more similar they are. In other words, we define higher-order correlations in a hypergraph as “commonality,” which is the property of sharing a common group between two nodes. Our assumption is confirmed in Figure \ref{fig2}, where we see that the cosine similarity of two node features increases as the number of hyperedges shared by the two nodes increases. For example, two people belonging to the same group could mean that they have the same disposition, or they could belong to the same group and have similar dispositions. Therefore, the assumption of common group membership is important to exploit the high-dimensional correlations in the hypergraph. Also, the model implements group unit contrastive learning to efficiently learn homophily trends within the hypergraph. In previous HGCL, contrastive learning is based on nodes; however, we use group-based contrastive learning. Thus, nodes that belong to the same group are able to contrasted multiple times. 


Our proposed HyFi method enhances the classification of weak positive relations in shared group connections between nodes in a hypergraph. This approach generates higher quality embeddings than those produced by traditional hypergraph contrastive learning (HGCL) models. As a result, HyFi outperformed 10 supervised and 6 unsupervised baselines in node classification tasks across 10 datasets. It also outperformed both types of baselines in 10 separate node classification tasks. Furthermore, because HyFi does not modify the topology of the graph, it is widely applicable to a variety of datasets and is more efficient in terms of both time and memory usage compared to traditional HGCL models.

The main contributions of this paper are highlighted as follows:
\begin{itemize}
    \item We propose a novel  contrastive learning method, called HyFi. The method exploits the higher-order correlations in hypergraphs represented by commonality to achieve efficient contrast learning that is less dependent on graph augmentation than existing methods.
    \item We introduce a new relation called weak positive pair, which shows that setting the positive sample to be fine-grained is key to contrastive learning.
    \item We propose an augmentation scheme that adds noise to the node features in a way that does not preserve the topology of the hypergraph, and show that it is sufficient to generate positive samples that are comparable to various forms of augmentation.
\end{itemize}


\section{Related Work}
\label{sec:relatedworks}
\subsection{Hypergraph neural networks}

Hypergraph neural networks (HGNNs) extend graph neural networks(GNNs) by enabling complex interactions through hyperedges. Early models like HGNN\cite{feng2019hypergraph} and HyperGCN\cite{yadati2019hypergcn} convert hypergraphs into graphs for processing, leading to the loss of higher-order relationships. HNHN\cite{dong2020hnhn} improves this problem by using a two-step message passing that preserves these relationships. Attention-based models such as HCHA\cite{bai2021hypergraph} further refine node and hyperedge aggregations. Advanced frameworks such as  UniGNN\cite{huangunignn} and AllSet\cite{chien2021you} have generalized these concepts to hypergraphs, with UniGCNII focusing on preventing over-smoothing in deep networks. AllSet and its variant AllSetTransformer use sophisticated functions for improved aggregation. However, most models still struggle with fully capturing the nuanced interactions within hyperedges. Furthermore, a majority of these models are still primarily limited to the settings of supervised or semi-supervised learning.

\subsection{Graph contrastive learning}
The realm of graph contrastive learning has been largely inspired by its success in computer vision and natural language processing. Techniques like DGI\cite{velivckovic2018deep}, InfoGraph\cite{sun2019infograph}, GRACE\cite{zhu2020deep}, and MVGRL\cite{hassani2020contrastive} have shown promising results in obtaining expressive representations for graphs and nodes by maximizing mutual information between various levels of graph structure. More recent approaches like GCA\cite{zhu2021graph} and DGCL\cite{zhao2021graph} have introduced novel methods for data augmentation, which is crucial for contrastive learning. However, these methods often require manual trial-and-error, cumbersome search processes, or expensive domain knowledge for augmentation selection, limiting their scalability and general applicability.

\subsection{Hypergraph contrastive learning}
Contrastive learning on hypergraphs, though still in its nascent stages, is beginning to receive more attention. Initial studies such as S\textsuperscript{2}-HHGR\cite{zhang2021double} have focused on applying contrastive learning for specific applications like group recommendation, employing unique hypergraph augmentation schemes. However, they often overlook the broader potential of hypergraph structures in general representation learning. Recent efforts in this area are starting to explore more comprehensive approaches that fully leverage the complex structure of hypergraphs. For instance, TriCL\cite{lee2023m} uses a tri-directional contrast mechanism to learn node embeddings by contrasting node, group, and membership labels. However, many of these methods still focus on specific applications and do not fully exploit the rich, multi-relational nature of hypergraphs.

In this paper, we propose a hypergraph-specific contrastive learning method that exploits the high-order correlations of hypergraphs by presenting a fine-grained approach to learning node pairs. We also present a method that does not exploit the structure of hypergraphs, enabling more efficient learning than existing hypergraph models.

\section{Preliminary}
\label{sec:preliminary}
\subsection{Preliminaries and Notation}
In this section, we introduce essential concepts relevant to HyFi, and present preliminaries concerning hypergraphs and hypergraph neural networks. The notations utilized are concisely summarized in Table \ref{tab:notation}.

\subsubsection{Hypergraphs}
HyFi employs hypergraphs as a foundational structure, denoted by \( G = (V, E) \), where \( G \) represents the hypergraph, \( V \) the set of nodes, and \( E \) the set of hyperedges.  The nodes and hyperedges are denoted as \( V = \{v_1, v_2, \ldots, v_{|V|}\} \) and \( E = \{e_1, e_2, \ldots, e_{|E|}\} \), respectively. The node feature matrix is defined by \( X \in \mathbb{R}^{|V| \times d} \), where each node \( v_i \)'s feature vector \( x_i \) is represented by \( X[i, :]^T \in \mathbb{R}^d \). In general, a hypergraph can be alternatively characterized by its incidence matrix \( H \in \{0, 1\}^{|V| \times |E|} \), where the entries are defined as \( h_{ij} = 1 \) if node \( v_i \) is included in hyperedge \( e_j \), and \( h_{ij} = 0 \) otherwise.

\subsubsection{Hypergraph neural networks}

Prior studies on HGNNs have predominantly adopted a two-phase neighborhood aggregation approach that includes both node-to-hyperedge and hyperedge-to-node aggregations. These networks use an iterative process to update hyperedge representations by accumulating the representations of neighboring nodes, and similarly, to update node representations by accumulating those of neighboring hyperedges. At the k-th layer, the representations of nodes and hyperedges are denoted as \( P^{(k)} \in \mathbb{R}^{|V| \times d^\prime_k} \) and \( Q^{(k)} \in \mathbb{R}^{|E| \times d^{\prime\prime}_k} \), respectively. More precisely, the k-th layer function in a Hypergraph neural network can be expressed as \( q_j^{(k)} = f_{V \rightarrow E}^{(k)}(q_{j}^{(k-1)}, p_{i}^{(k-1)} : v_i \in e_j) \) and \( p_i^{(k)} = f_{E \rightarrow V}^{(k)}(p_{i}^{(k-1)}, q_{j}^{(k)} : v_i \in e_j) \), where \( p_i^{(0)} = x_i \). Therefore, the choice of aggregation functions \( f_{V \rightarrow E}(\cdot) \) and \( f_{E \rightarrow V}(\cdot) \) is critical to the performance of the model. We will use HGNN model\cite{feng2019hypergraph} as the backbone of the HyFi encoder.

\begin{table}[t!]
\centering
\caption{Primary Notations in this paper.}
\label{tab:notation}
\resizebox{\columnwidth}{!}{
\renewcommand{\arraystretch}{1.2}
\begin{tabular}{cl}
\hline
\textbf{Notation} & \textbf{Definition} \\
\hline
$G = (V, E)$ & Hypergraph with nodes $V$ and hyperedges $E$ \\
$H$ & Incidence matrix \\
$X$ & Node feature matrix \\
$V$ & Hypergraph with nodes, where $V = \{v_1, v_2, \ldots, v_{|V|}\}$ \\
$E$ & Hypergraph with hyperedges, where $E = \{e_1, e_2, \ldots, v_{|E|}\}$ \\
$P^{(l)}$ & Node embedding matrix at layer $l$\\
$Q^{(l)}$ & Hyperedge embedding matrix at layer $l$\\
$l$ & Number of layers\\
$D_v$ & Diagonal matrix, Degree of node\\
$D_e$ & Diagonal matrix, Degree of hyperedge\\
$W$ & Diagonal matrix, Weight of hyperedge\\
$\mathcal{N}(0, \sigma^2)$ & noises from Gaussian distribution with 0 mean and variance $\sigma^2$\\
$f_\theta(\cdot)$ & HGNN Encoder\\  
$g_\pi(\cdot)$ & Node embedding projection head\\
$g_\psi(\cdot)$ & Hyperedge embedding projection head\\
$Z$ & Final Node embedding after passing the projection head\\
$Y$ & Final Hyperedge embedding after passing the projection head\\
$sim(u, v)$ & Cosine similarity $\mathbf{u}^\mathsf{T} \mathbf{v} / \|\mathbf{u}\| \|\mathbf{v}\|$\\
$N_{v_i}$ & Set of other nodes in the group to which node $v_i$ belongs\\
$M$ & Number of Noise views\\
\hline
\end{tabular}}
\end{table}


\section{Proposed Method}
\label{sec:proposed}
\begin{figure*}[t!]
  \centering
  \includegraphics[width=0.90\textwidth, 
  ]{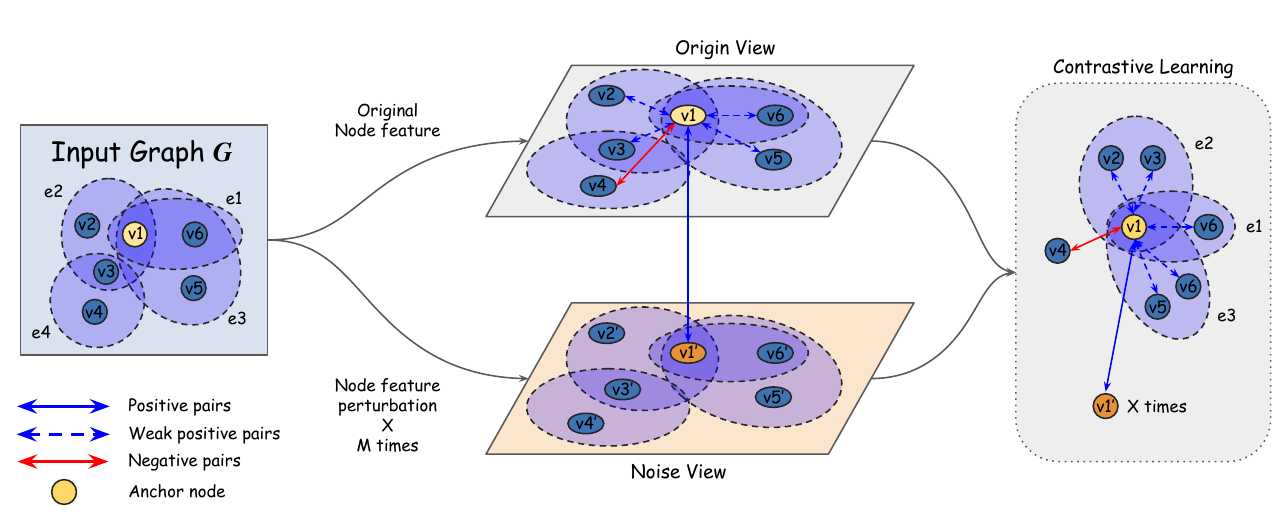}
  \caption{Illustration of the node-level contrastive loss calculation process in HyFi. HyFi uses both the origin view and the noise view to calculate contrastive loss. In the origin view, there is a negative and a weak positive pair, and in the noise view there is a positive pair, which is a node in the same position as the anchor node.}
  \label{fig:contrastloss}
\end{figure*}

In this section, we provide a detailed exposition of our proposed HyFi, the overall framework of which is illustrated in Fig. \ref{fig:contrastloss}.


\subsection{HyFi}

Here, we propose HyFi, a method that exploits the high-dimensional information of hypergraphs for fine-grained contrastive learning by introducing weak positive pairs that share the same group. Unlike existing methods in hypergraph contrastive learning, HyFi fully exploits the homophily feature of hypergraphs, which is based on group commonality, to achieve superior performance. In addition, it avoids augmentations that modify the topology of the hypergraph. Instead, HyFi adds Gaussian random noise to the node features, allowing for fast training speeds, efficient memory usage, and robustness against constraints typically imposed by graph augmentation.


\subsection{HyFi: Node perturbation}


When we have a hypergraph \( G = (V, E) \) along with node features \( X \), our approach involves perturbing only \( X \) without altering the topology of the hypergraph. The perturbation is executed by adding Gaussian-distributed random noise to the original node features. Since the information values of the Node feature are mostly one-hot vectors or normalized between 0 and 1, we perturb the values so that they are between 0 and 1. Our proposed method to perturb the node feature $X$ can be described as,
\begin{equation}
\label{equ:add_noise1}
    X^{\prime} = X + (-1)^X \cdot |\epsilon|, \quad \epsilon \sim \mathcal{N}(0, \sigma^2)
\end{equation},
where $X$ and $X^{\prime}$ are the original node feature and the node feature with noise added, respectively. $\epsilon$ denotes noise with a Gaussian distribution with a mean of zero and a variance of $\sigma^2$. 


\subsection{HyFi: Hypergraph encoder}

In our framework, we employ an HGNN model for the hypergraph encoder, denoted by \( f_\theta(\cdot) \). As input to the encoder, we take the same hypergraph \( G = (V, E)\), but different original node features $X$ and node features with noise added $X^{\prime}$. We denote the view generated with $X$ as the origin view, and the view generated with $X$ as the noise view. The encoder generates embeddings for the nodes and hyperedges of the hypergraph, respectively. The representation of this process is as follows:

\begin{equation}
\begin{aligned}
    f_\theta(G, X) &= (P, Q),  \\
    f_\theta(G, X^{\prime}) &= (P^{\prime}, Q^{\prime}),
\end{aligned}
\end{equation},
where \( P \) and \( Q \) represent the embeddings of nodes and hyperedges from the original view, respectively, while \( P^{\prime} \) and \( Q^{\prime} \) correspond to the embeddings from the noise view. 

\begin{equation}
\begin{aligned}
    Q^{(k)} &= \sigma\left( D_E^{-1} H^T D_V^{-1/2} P^{(k-1)} \Theta_E^{(k)} + b_E^{(k)} \right), \\
    P^{(k)} &= \sigma\left( D_V^{-1/2} HW Q^{(k)} \Theta_V^{(k)} + b_V^{(k)} \right),
\end{aligned}
\end{equation},
where \( P^{(0)} = X \) and \( W \) is the identity matrix. 

The encoder generates embeddings for both the nodes and hyperedges of the hypergraph, and we incorporate two projection heads, symbolized by \( g_{\phi}(\cdot)\) and \( g_{\psi}(\cdot)\), to map the embeddings of nodes and hyperedges, respectively. These projection heads are built using a two-layer MLP with ELU activation function\cite{clevert2015fast} as, 
\begin{equation}
\label{equ:add_noise2}
\begin{split}
    Z & := g_{\phi}(P)  \quad Z^{\prime} := g_{\phi}(P^{\prime}), \\
    Y & := g_{\psi}(Q)  \quad Y^{\prime} := g_{\psi}(Q^{\prime}),
\end{split}
\end{equation},
where \(\prime\) is the embedding of the noise view from the perturbed node features.




\subsection{HyFi: Contrastive Loss}

HyFi computes contrastive loss at both the node and edge levels. The core principle of this computation is that nodes sharing hyperedges are considered weak positive pairs in node-level contrastive learning. This assumption is based on the premise that nodes within the same group are likely to share similarities. Furthermore, we exploit the unique perspective of each group by performing the contrastive loss computation specifically within each group. This method allows us to focus on a node's group affiliation, allowing us to compute the contrastive loss proportional to the number of hyperedges shared between two nodes. We call this method group-unit contrastive learning. Computing contrastive loss at the edge level uses information from the nodes that belong to each hyperedge; therefore, the computational process is the same at the node level and at the edge level.


\subsubsection{Node-level contrastive loss}

We consider a view with the original node features, \(X\), as the original view and the noisy node features, \(X^{\prime}\), as the noisy view. When calculating contrastive loss at the node level, we consider three types of pairs: positive pairs, weakly positive pairs, and negative pairs. The number of hyperedges shared between each node can be determined as the matrix product of the incidence matrix \(H\) and its transpose \(H^T\). Here, the diagonal elements of the resulting matrix represent the number of hyperedges each node belongs to, while the off-diagonal elements represent the number of hyperedges shared between nodes \(i\) and \(j\). Mathematically, this can be expressed as:

\[ E = H \cdot H^T \]

where \(E\) is the resulting matrix. In \(E\), the element \(E_{ii}\) represents the number of hyperedges node \(i\) belongs to, and the element \(E_{ij}\) (for \(i \neq j\)) represents the number of hyperedges shared between nodes \(i\) and \(j\).

Nodes that share the same hyperedge in the original view are considered weak positive pairs. All other nodes, which do not share a hyperedge with the anchor node, are categorized as negative pairs. For example, if a weak positive pair node represents a relationship within the same group as me, then a node that belongs to a group I have never interacted with is defined as a negative pair. Furthermore, we utilize the noise view embedding—obtained by perturbing the node features earlier—as a positive pair. Consequently, the number of positive pairs equals the number of noise views.


We also assign a weight to the weak positive pair, which represents the probability that a node is a positive pair with the anchor node. This weight is calculated as the ratio of the number of shared hyperedges between the two nodes to the total number of hyperedges of the anchor node. For instance, if the anchor node is part of 3 hyperedges and shares 2 hyperedges with another node, the weight would be 2/3. The contrast loss is also calculated based on the number of shared hyperedges, allowing us to compute the loss on a group basis. Therefore, the final weight for the last weak positive pair is determined by multiplying the number of shared hyperedges by the ratio of the number of shared hyperedges identified earlier. The formula for the weights of weak positive is as follows:


\begin{equation}
\begin{split}
    w_{i,j}^{pos} &= \underbrace{|E_{i,j}|}_{\text{for group unit contrast}} \cdot \underbrace{\frac{|E_{i,j}|}{|E_{i,i}|}}_{\text{postivie probability}}, \\
\end{split}
\end{equation}

Here, $w_{i,j}^{pos}$ denotes the weak positive weight. The contrastive losses for $\mathcal{L}_n^{pos}$, $\mathcal{L}_n^{weak\_pos}$, and $\mathcal{L}_n^{neg}$ are calculated as follows:

\begin{equation}
\begin{split}
&\mathcal{L}_n^{pos} = \sum_{m \in M} e^{\text{sim}(z_i, z_{m,i}^{\prime})/\tau_n}, \\
&\mathcal{L}_n^{weak\_pos} = \sum_{j \in N_{z_i}} w_{i,j}^{pos} \cdot e^{\text{sim}(z_i, z_j)}/{\tau_n}, \\
&\mathcal{L}_n^{neg} = \sum_{j \notin N_{z_i}} e^{\text{sim}(z_i, z_j)}/{\tau_n}, \\
\end{split}
\end{equation},
where $\tau_n$ is a temperature parameter, $M$ is set of noise views.

With the Origin View at the anchor $z_i$ and the set of Noise View as $z_{m}^{\prime}$. The node-level contrastive loss $\mathcal{L}_n(z_i, z_{m}^{\prime})$ is defined as follows:

\begin{equation}
    \mathcal{L}_n(z_i, z_{m}^{\prime}) = -\log \left( \frac{\mathcal{L}_n^{pos} + \mathcal{L}_n^{weak\_pos}}{\mathcal{L}_n^{pos} + \mathcal{L}_n^{weak\_pos} + \mathcal{L}_n^{neg}} \right)
\end{equation}
.

\subsubsection{Edge-level contrastive loss}

The calculation of edge-level contrastive loss is the same process as the calculation of node-level contrastive loss. However, at the edge level, the node and hyperedge relationships at the node level are reversed, i.e., at the hyperedge level, the shared nodes between hyperedges are utilized instead of the shared hyperedges between each node. 
\subsubsection{Final contrastive loss}
Finally, the final loss $\mathcal{L}$ is calculated by summing the node-Level contrastive loss $loss_n$ and the edge-level contrastive loss $loss_e$ as follows:

\begin{equation}
\begin{aligned}
    \mathcal{L} = \sum_{i=1}^V \mathcal{L}_n(z_i, z_{m}^{\prime}) + \alpha \cdot \sum_{i=1}^E \mathcal{L}_e(y_i, y_{m}^{\prime})
\end{aligned}
\end{equation},
where $\alpha$ are the weights for the edge-level loss.

\subsection{Complexity Analysis}


The training algorithm for HyFi is thoroughly outlined in Algorithm 1. Additionally, we conducted an analysis of the time and space complexity of the algorithm.

\begin{algorithm}
    \caption{HyFi}
    \begin{algorithmic}[1]
        \State \textbf{Input:} Hypergraph $G = (V, E)$, Node Feature $X$
        \State \textbf{Parameters:} weight for edge loss $\alpha$, weight matrix for node $W_v$, weight matrix for edge $W_e$, standard deviation of Gaussian noise $\sigma^2$, number of noise views $m$
        \State \textbf{Output:} Trained HGNN encoder $\mathit{f}_{\Theta}$, node embeddings $Q$, edge embeddings $P$
        \For{$epoch \gets 1,2,\ldots$}
            \State $Q, P \gets \mathit{f}_{\Theta}(G, X)$
            \State $Q_{proj}, P_{proj} \gets \mathit{g}_{\Phi}(Q), \mathit{g}_{\Psi}(P)$
            \For{$i \gets 1 \text{ to } m$}
                \State $X' \gets X + \mathcal{N}(0, \sigma^2)$
                \State $Q', P' \gets \mathit{f}_{\Theta}(G, X')$
                \State $Q'_{proj}, P'_{proj} \gets \mathit{g}_{\Phi}(Q'), \mathit{g}_{\Psi}(P')$ 
            \EndFor
            \State $\mathcal{L}_{node} \gets Loss(Q_{proj}, Q'_{proj}, W_v)$
            \State $\mathcal{L}_{edge} \gets Loss(P_{proj}, P'_{proj}, W_e)$
            \State $\mathcal{L} \gets \mathcal{L}_{node} + \alpha \cdot \mathcal{L}_{edge}$
            \State Perform gradient descent on $\Theta$ to minimize $\mathcal{L}$
        \EndFor
    \end{algorithmic}
\end{algorithm}

\subsubsection{Time Complexity}

We assess time complexity based on the pseudocode. Initially, the generation of noise features in line 8 has a time complexity of \( O(m \cdot n_d \cdot |V|) \), where \( n_d \) is the dimensionality of the node features, \( |V| \) represents the number of nodes, and \( m \) is the number of noise views. For the computation of the node-level contrastive loss in line 12, the cosine similarity for each node is calculated. This includes computing the similarity within the original view and between the original view and the noise view. Cosine similarity calculation has a complexity of \( O(n_d) \) per node. When computed for all nodes and for each noise view, the total time complexity of node-level contrastive loss is \( O(m \cdot |V| \cdot n_d) \). Similarly, the computation of the edge-level contrastive loss follows the node-level process but applies to the edge embeddings \( P \), resulting in a time complexity of \( O(m \cdot |E| \cdot n_d) \), where \( |E| \) denotes the total number of hyperedges. Therefore, the time complexity per training iteration is \( O(m \cdot n_d \cdot (|V| + |E|)) \). All calculations for HyFi can be efficiently performed on a GPU.

\subsubsection{Space Complexity}

The graph consists of a set of nodes \( V \) and a set of edges \( E \). Depending on the representation of the graph, which could be an adjacency list or matrix, the space required is typically \( O(|V| + |E|) \). Node features occupy a space complexity of \( O(|V| \cdot n_d) \) for a \( n_d \)-dimensional feature vector per node. Node and edge embeddings, which correspond to vectors for each node and edge, occupy \( O(|V| \cdot n_d) \) and \( O(|E| \cdot n_d) \), respectively. 

The weight matrices for nodes and edges, necessary for calculating the weights for weak positive and weak negative pairs, require a space of \( O(|V| \cdot n_d) \) and \( O(|E| \cdot n_d) \), respectively. The new feature sets generated by adding noise to each node feature match the size of the original features, contributing an additional \( O(m \cdot |V| \cdot n_d) \) to the space complexity, where \( n \) is the number of noise views. Thus, the total space complexity of the algorithm is the sum of these components.

When considering the most significant terms in the space complexity, the weight matrices for nodes and edges are \( O(|V| \cdot n_d) \) and \( O(|E| \cdot n_d) \), respectively. HyFi maintains the original graph topology, only augmenting the node feature \( X \) according to the noise view, facilitating more efficient memory usage compared to models that augment the graph topology itself.

\section{Experiments}
\label{sec:experiments}
In this section, we conduct experiments to evaluate HyFi through answering the following research questions.
\begin{itemize}
    \item \textbf{RQ1 (Generalizability)}: How does HyFi outperform baselines on node classification tasks?
    \item \textbf{RQ2 (Efficiency)}: How efficient is HyFi regarding GPU memory usage and training time?
    \item \textbf{RQ3 (Ablation Study)}: How do the newly introduced modules in HyFi affect its overall performance?
    \item \textbf{RQ4 (Type of Augmentation)}: How do different augmentation methods affect performance?
    \item \textbf{RQ5 (Effect of Positive sample)}: How does the number of positive samples affect hypergraph contrastive learning?
\end{itemize}

\begin{table}[t!]
\centering
\caption{Statistics for datasets.}
\begin{tabular}{lccc}
\hline
Dataset & \# Nodes & \# Hyperedges & \# Classes \\
\hline \hline
Cora-C & 1,434 & 1,579 & 7 \\
Citeseer & 1,458 & 1,079 & 6 \\
Pubmed & 3,840 & 7,963 & 3 \\
Cora-A & 2,388 & 1,072 & 7 \\
DBLP & 41,302 & 22,363 & 6 \\
Zoo & 101 & 43 & 7 \\
20News & 16,242 & 100 & 4 \\
Mushroom & 8,124 & 298 & 2 \\
NTU2012 & 2,012 & 2,012 & 67 \\
ModelNet40 & 12,311 & 12,311 & 40 \\
\hline
\end{tabular}
\label{tab:dataset}
\end{table}

\subsection{Experimental Setup}

\subsubsection{Datasets}

To compare the performance of HyFi, we use a total of 10 datasets that can be categorized into: (1) citation network datasets (Cora-C, Citeseer, and Pubmed \cite{sen2008collective}), (2) co-authorship network datasets (Cora-A and DBLP \cite{rossi2015network}), (3) computer vision and graphics datasets (NTU2012 \cite{chen2003visual} and ModelNet40 \cite{wu20153d}), and (4) datasets from the UCI Categorical Machine Learning Repository (Zoo, 20Newsgroups, and Mushroom \cite{asuncion2007ucireferences}). The statistics of these datasets are provided in Table \ref{tab:dataset}.

\subsubsection{Hyperparameter Setting}
For all datasets, we used AdamW as the optimizer\cite{loshchilov2018decoupled}. To perturb the node features, we use random noise from a Gaussian distribution with mean zero and standard deviation $\sigma^2$ as a hyperparameter. The number of positive samples equals the number of noise views, which is also a hyperparameter.

\begin{table*}[t!]
\centering
\caption{Comparison of node classification accuracy. The highest scores in each dataset are highlighted in bold, indicating the best performance by method, while the second highest scores are underlined. The `A.R.' indicates the average rank across all datasets. `-' indicates that the results are not available in published papers.}
\resizebox{\textwidth}{!}{
\renewcommand{\arraystretch}{1.5}
\begin{tabular}{lcccccccccccc}
\hline
& Method & {Cora-C} & {Citeseer} & {Pubmed} & {Cora-A} & {DBLP} & {Zoo} & {20News} & {Mushroom} & {NTU2012} & {ModelNet40} & {A.R.$\downarrow$}\\
\hline
\hline
\multirow{10}{*}{\rotatebox[origin=c]{90}{Supervised}}
& MLP & 60.32\textpm1.5 & 62.06\textpm2.3 & 76.27\textpm1.1 & 64.05\textpm1.4 & 81.18\textpm0.2 & 75.62\textpm9.5 & 79.19\textpm0.5 & 99.58\textpm0.3 & 65.17\textpm2.3 & 93.75\textpm0.6 & 11.4 \\ 
& GCN & 77.11\textpm1.8 & 66.07\textpm2.4 & 82.63\textpm0.6 & 73.66\textpm1.3 & 87.58\textpm0.2 & 36.79\textpm9.6 & - & 92.47\textpm0.9 & 71.17\textpm2.4 & 91.67\textpm0.2 & 10.7 \\ 
& GAT & 77.75\textpm2.1 & 67.62\textpm2.5 & 81.96\textpm0.7 & 74.52\textpm1.3 & 88.59\textpm0.1 & 36.48\textpm10.0 & - & - & 70.94\textpm2.6 & 91.43\textpm0.3 & 10.0 \\ 
& HGNN & 77.50\textpm1.8 & 66.16\textpm2.3 & 83.52\textpm0.7 & 74.38\textpm1.2 & 88.32\textpm0.3 & 78.58\textpm11.1 & \underline{80.15\textpm0.3} & 98.59\textpm0.5 & 72.03\textpm2.4 & 92.23\textpm0.2 & 7.1 \\
& HyperConv & 76.19\textpm2.1 & 64.12\textpm2.6 & 83.42\textpm0.6 & 73.52\textpm1.0 & 88.83\textpm0.2 & 62.53\textpm14.5 & 79.83\textpm0.4 & 97.56\textpm0.6 & 72.62\textpm2.6 & 91.84\textpm0.1 & 8.7 \\
& HNHN & 76.21\textpm1.7 & 67.28\textpm2.2 & 80.97\textpm0.9 & 74.88\textpm1.6 & 86.71\textpm1.2 & 78.89\textpm10.2 & 79.51\textpm0.4 & \underline{99.78\textpm0.1} & 71.45\textpm3.2 & 92.96\textpm0.2 & 7.9 \\
& HyperGCN & 64.11\textpm7.4 & 59.92\textpm9.6 & 78.40\textpm9.2 & 60.65\textpm9.2 & 76.59\textpm7.6 & 40.86\textpm2.1 & 77.31\textpm6.0 & 48.26\textpm0.3 & 46.05\textpm3.9 & 69.23\textpm2.8 & 14.1 \\
& HyperSAGE & 64.98\textpm5.3 & 52.43\textpm9.4 & 79.49\textpm8.7 & 64.59\textpm4.3 & 79.63\textpm8.6 & 40.86\textpm2.1 & - & - & - & - & 13.7 \\
& UniGCN & 77.91\textpm1.9 & 66.40\textpm1.9 & \textbf{84.08\textpm0.7} & 77.30\textpm1.4 & 90.31\textpm0.2 & 72.10\textpm12.1 & \textbf{80.24\textpm0.4} & 98.84\textpm0.5 & 73.27\textpm2.7 & 94.62\textpm0.2 & 5.0 \\
& AllSet & 76.21\textpm1.7 & 67.83\textpm1.8 & 82.85\textpm0.9 & 76.94\textpm1.3 & 90.07\textpm0.3 & 72.72\textpm11.8 & 79.90\textpm0.4 & \underline{99.78\textpm0.1} & \underline{75.09\textpm2.5} & 96.85\textpm0.2 & 5.0\\
\hline
\multirow{7}{*}{\rotatebox[origin=c]{90}{Unsupervised}}
& Random-Init & 63.62\textpm3.1 & 60.44\textpm2.5 & 67.49\textpm2.2 & 66.27\textpm2.2 & 76.57\textpm0.6 & 78.43\textpm11.0 & 77.14\textpm0.6 & 97.40\textpm0.6 & 74.39\textpm2.6 & 96.29\textpm0.3 & 10.9 \\
& Node2vec & 70.99\textpm1.4 & 53.85\textpm1.9 & 78.75\textpm0.9 & 58.50\textpm2.1 & 72.09\textpm0.3 & 17.02\textpm4.1 & 63.35\textpm1.7 & 88.16\textpm0.8 & 67.72\textpm2.1 & 84.94\textpm0.4 & 14.6 \\
& DGI & 78.17\textpm1.4 & 68.81\textpm1.8 & 80.83\textpm0.6 & 76.94\textpm1.1 & 88.00\textpm0.2 & 36.54\textpm9.7 & - & - & 72.01\textpm2.5 & 92.18\textpm0.2 & 8.2\\
& GRACE & 79.11\textpm1.7 & 68.65\textpm1.7 & 80.08\textpm0.7 & 76.59\textpm1.0 & - & 37.07\textpm9.3 & - & - & 70.51\textpm2.4 & 90.68\textpm0.3 & 9.4 \\
& S\textsuperscript{2}-HHGR & 78.08\textpm1.7 & 68.21\textpm1.8 & 82.13\textpm0.6 & 78.15\textpm1.1 & 88.69\textpm0.2 & \textbf{80.06\textpm11.1} & 79.75\textpm0.3 & 97.15\textpm0.5 & 73.95\textpm2.4 & 93.26\textpm0.2 & 5.7\\
& TriCL & \underline{81.03\textpm1.3} & \underline{71.97\textpm1.3} & 83.80\textpm0.6 & \textbf{82.22\textpm1.1} & \textbf{90.93\textpm0.2} & 79.47\textpm11.0 & 79.93\textpm0.2 & 98.93\textpm0.3 & 74.63\textpm2.5 & \underline{97.33\textpm0.1} & \underline{2.5} \\
& \textbf{HyFi} & \textbf{81.48\textpm1.5} & \textbf{72.57\textpm1.1} & \underline{83.82\textpm0.6} & \underline{79.33\textpm1.2} & \underline{90.41\textpm0.2} & \underline{80.02\textpm10.9} & 79.76\textpm0.3 & \textbf{99.79\textpm0.2} & \textbf{75.10\textpm2.6} & \textbf{97.38\textpm0.1} & \textbf{1.9} \\
\hline
\end{tabular}}
\label{table:node_classification}
\end{table*}

\subsubsection{Compared baselines}

HyFi was compared against 10 (semi-)supervised models and 5 unsupervised models. Given that graph-based methods are not directly applicable to hypergraphs, hypergraphs were transformed into graphs through clique expansion before applying these methods. We focused particularly on comparisons with the state-of-the-art model in HGCL, which is TriCL. To ensure a fair comparison, we assume the use of the same encoder (HGNN) architecture. For the baselines other than TriCL, results from their respective references are cited. 


\subsubsection{Evaluation protocol}
For the node classification task, we follow the standard linear evaluation protocol introduced in \cite{velivckovic2018deep}. First, an encoder is trained in a fully unsupervised manner to generate node representations. A simple linear classifier is then trained on top of these fixed representations without backpropagation the gradient through the encoder. For all datasets, nodes are randomly assigned to the training, validation, and test sets in proportions of 10\%, 10\%, and 80\%, respectively, following the approach in \cite{zhu2020deep} and \cite{thakoorlarge}. In the unsupervised setting, we evaluated the model using 20 dataset splits with 5 random weight initializations and reported the average accuracy for each dataset. In the supervised setting, 20 dataset splits were used, each with a different model initialization, and the average accuracy was reported. The experiment was repeated a total of five times, and the average results are reported here.

\subsection{Generalizability (RQ1)}

To evaluate the overall performance of HyFi, we performed comparative node classification experiments using different baseline models. As shown in the results in Table \ref{table:node_classification}, the HyFi model performed well on the node classification task in both supervised and unsupervised models. HyFi consistently ranked among the top performers on a variety of datasets and had the highest average ranking. Notably, it outperformed TriCL on average ranking, which is encouraging because HyFi outperforms TriCL while still using resources efficiently. These results highlight the importance of considering commonality, which is high-dimensional information in hypergraphs, and suggest that weak positive pair relationships are efficient for contrast learning. HyFi also has the advantage of being applicable to a wide variety of graphs because it relies little on graph augmentation. In fact, it ranks in the top 2 on almost all of the 9 datasets except 20News, and even on 20News, the performance gap to the top is very small.


\begin{table}[t!]
\centering
\caption{Comparisons of training time.}
\label{tab:speed}
\begin{tabular}{lcc}
\hline
Dataset & TriCL & HyFi \\
\hline
\hline
Cora-C & 21.79 s ($\times$1.71) & 12.71 s ($\times$1.00)\\
Citeseer & 58.91 s ($\times$3.17) & 18.60 s ($\times$1.00)\\
Pubmed & 350.68 s ($\times$5.07) & 69.22 s ($\times$1.00)\\
Cora-A & 33.82 s ($\times$1.40) & 24.21 s ($\times$1.00)\\
DBLP & 1145.62 s ($\times$2.38) & 481.86 s ($\times$1.00)\\
Zoo & 5.11 s ($\times$2.61) & 1.96 s ($\times$1.00)\\
20News & 286.86 s ($\times$5.68) & 50.47 s ($\times$1.00)\\
Mushroom & 251.69 s ($\times$7.95) & 31.67 s ($\times$1.00)\\
NTU2012 & 44.08 s ($\times$4.98) & 8.86 s ($\times$1.00)\\
ModelNet & 213.52 s ($\times$3.29) & 65.04 s ($\times$1.00)\\
\hline
\end{tabular}
\end{table}

\subsection{Efficiency (RQ2)}

To demonstrate the efficiency of the HyFi model, we conducted an experiment to compare it with another HGCL model, TriCL, in terms of training time and memory cost. To ensure a fair and consistent comparison, we set the hyperparameters between the two models as similar as possible. We experimented on the RTX A6000 (48GB GPU), and only employed batch processing in for the DBLP datasets, as not using batch processing would result in out-of-memory error.


\begin{figure*}[thbp!]
  \centering
  \begin{subfigure}[b]{0.2\textwidth}
    \includegraphics[width=\textwidth]{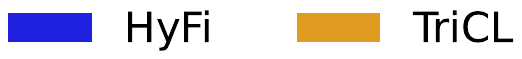}
  \end{subfigure}
  \\
  \begin{subfigure}[b]{0.19\textwidth}
    \includegraphics[width=\textwidth]{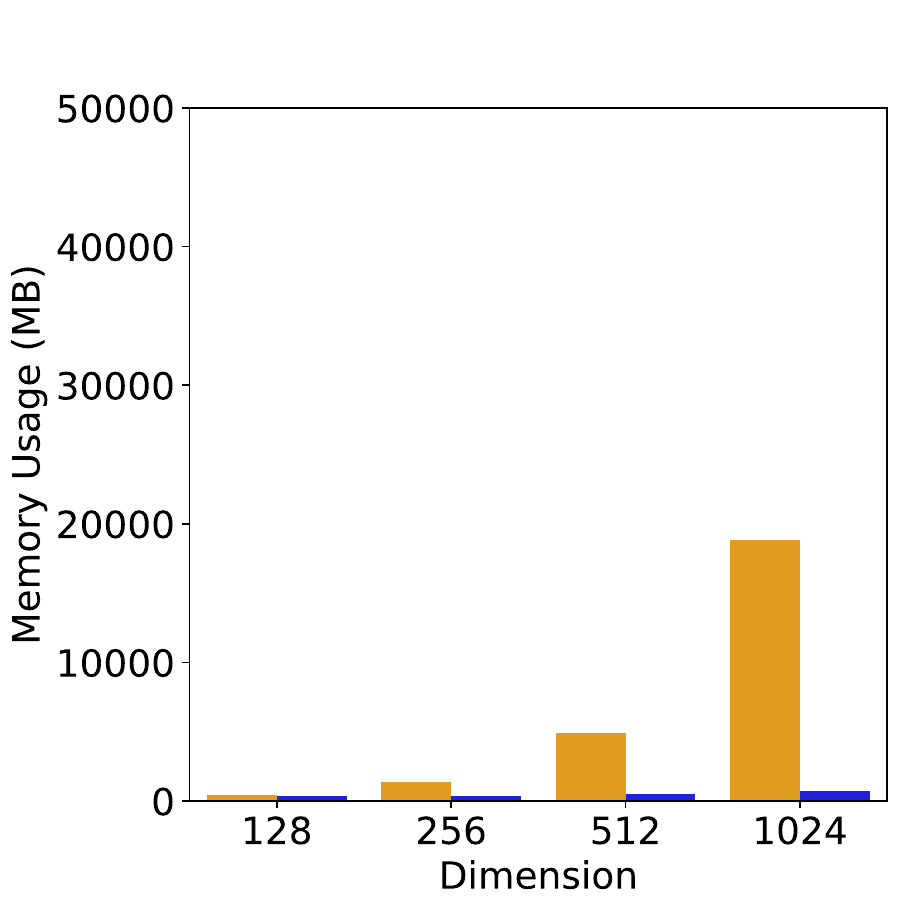}
    \caption{Cora-C}
  \end{subfigure}
  \begin{subfigure}[b]{0.19\textwidth}
    \includegraphics[width=\textwidth]{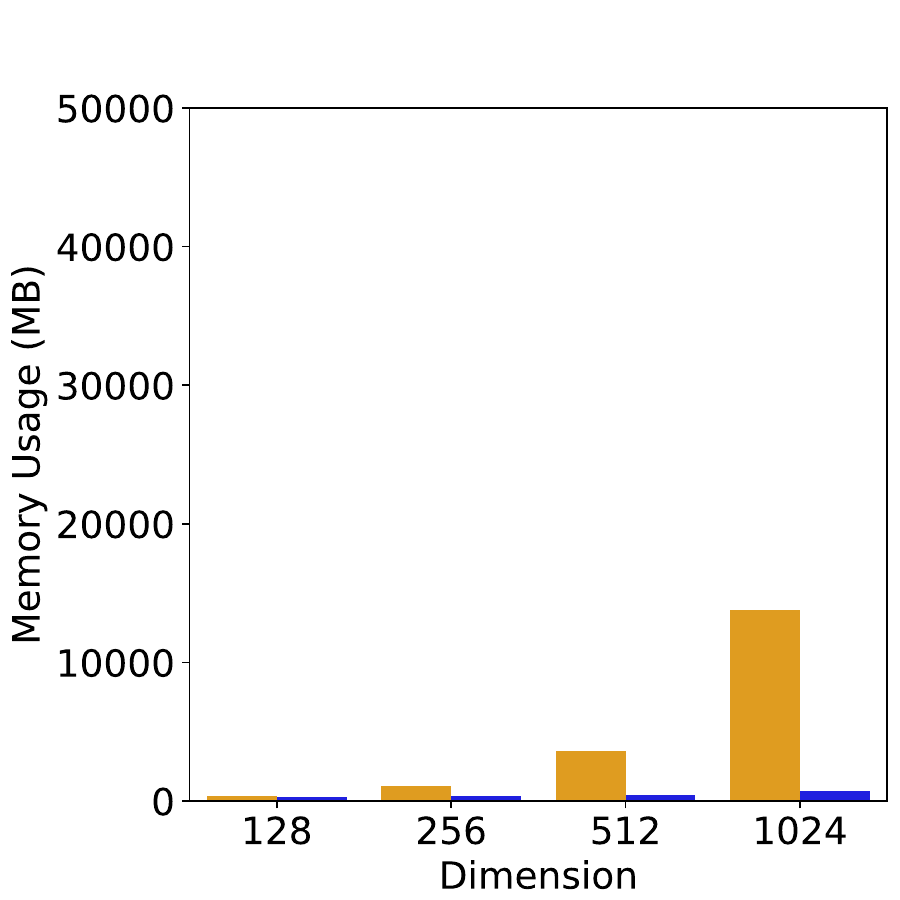}
    \caption{Citeseer}
  \end{subfigure}
  \begin{subfigure}[b]{0.19\textwidth}
    \includegraphics[width=\textwidth]{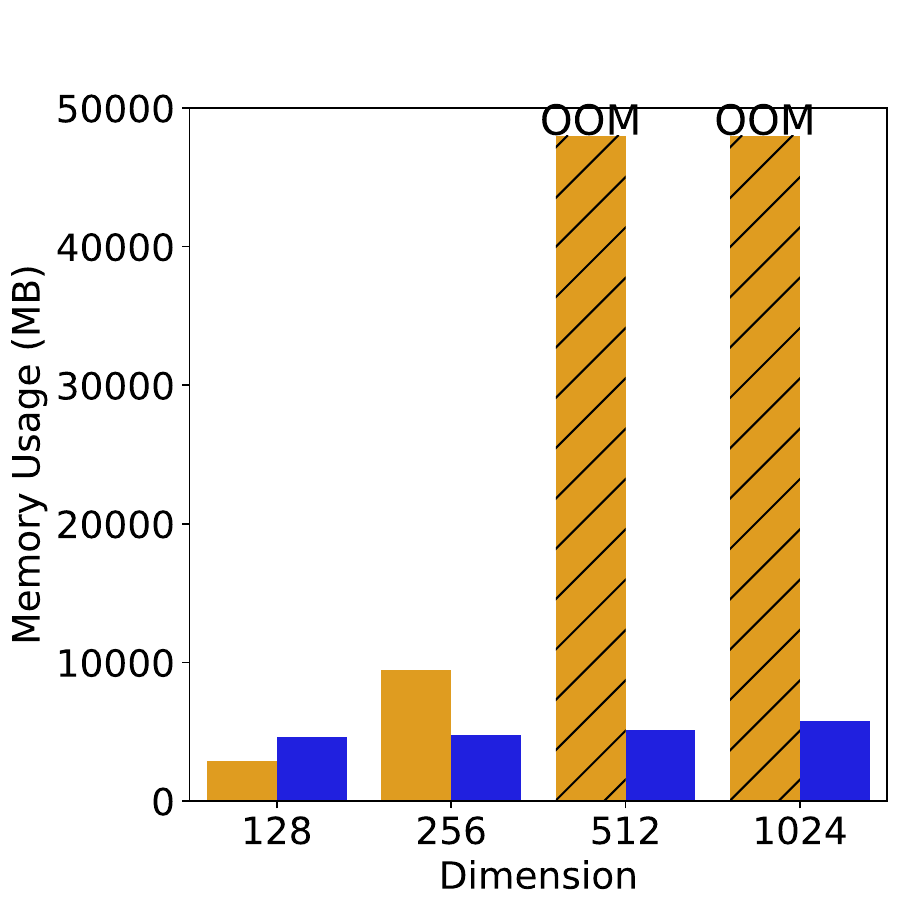}
    \caption{Pubmed}
  \end{subfigure}
  \begin{subfigure}[b]{0.19\textwidth}
    \includegraphics[width=\textwidth]{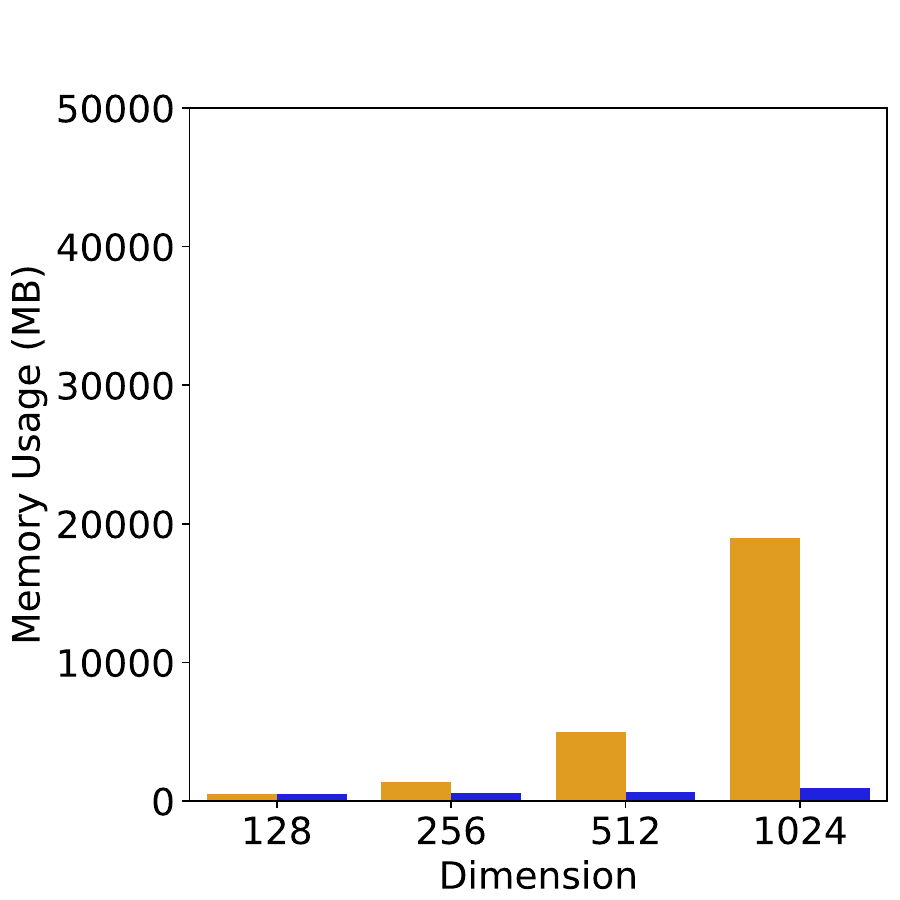}
    \caption{Cora-A}
  \end{subfigure}
  \begin{subfigure}[b]{0.19\textwidth}
    \includegraphics[width=\textwidth]{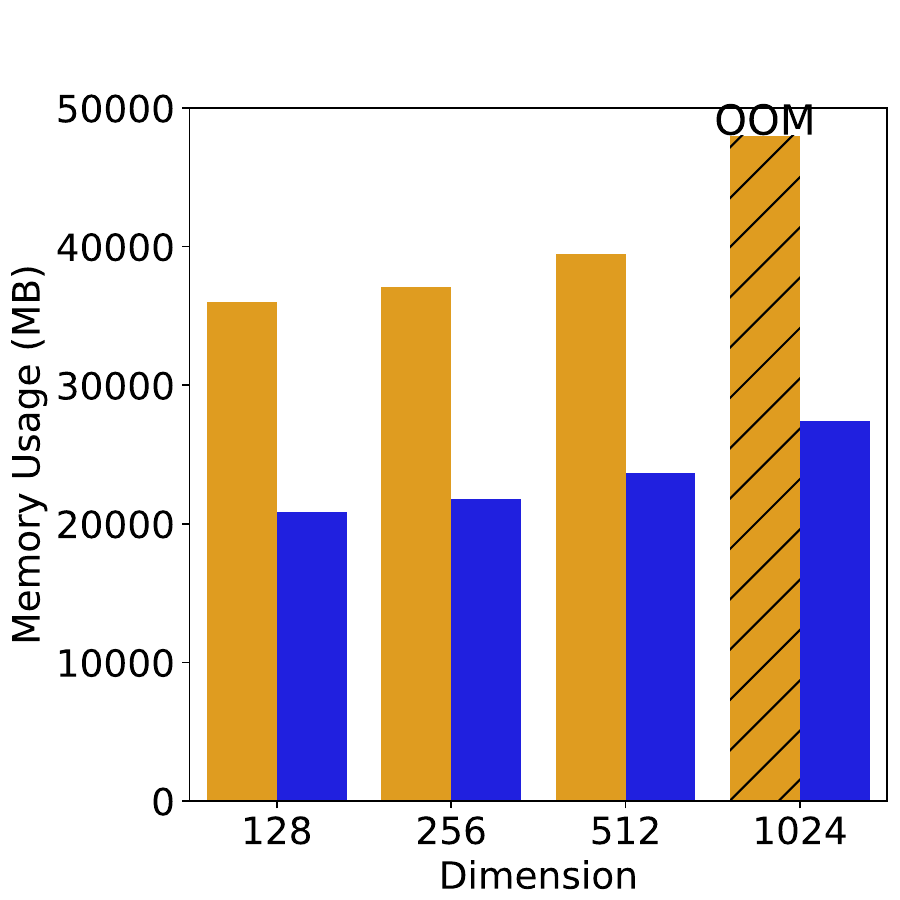}
    \caption{DBLP}
  \end{subfigure}
  \begin{subfigure}[b]{0.19\textwidth}
    \includegraphics[width=\textwidth]{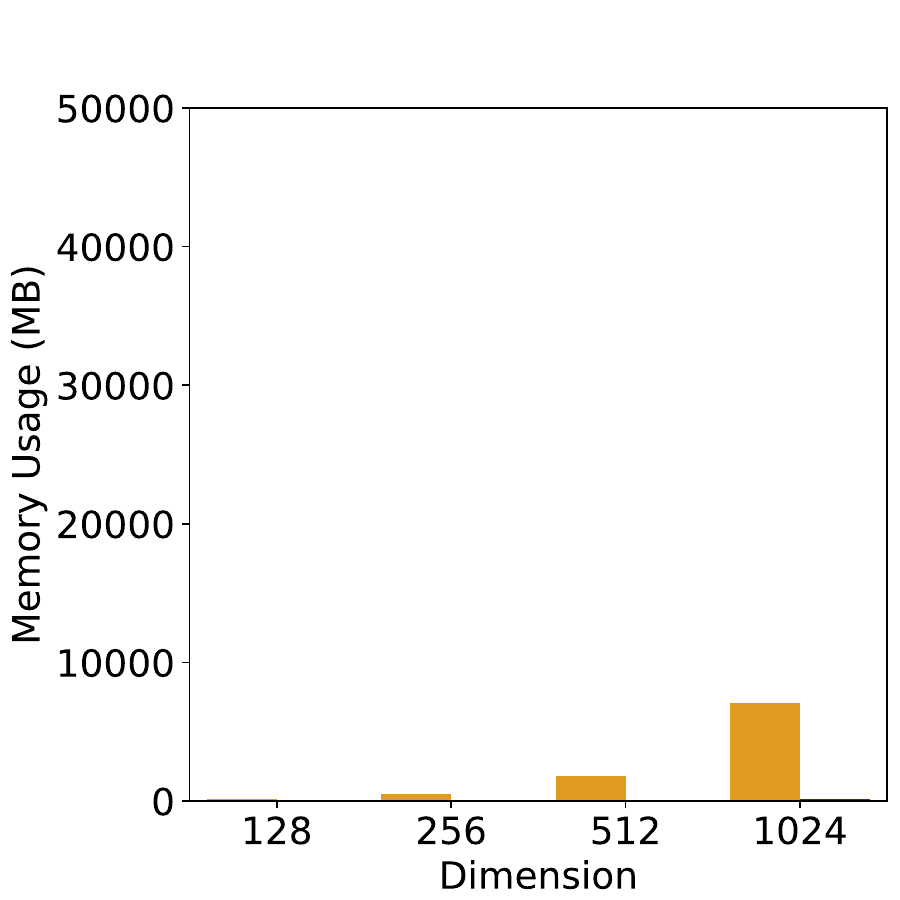}
    \caption{Zoo}
  \end{subfigure}
  \begin{subfigure}[b]{0.19\textwidth}
    \includegraphics[width=\textwidth]{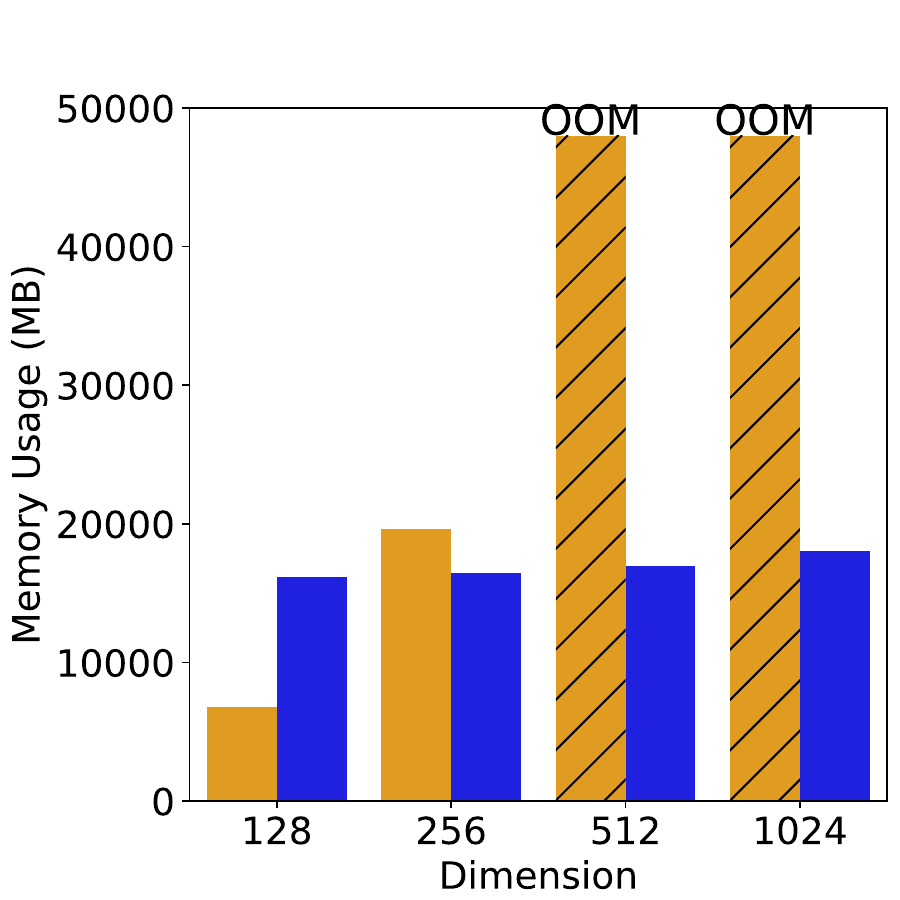}
    \caption{20News}
  \end{subfigure}
  \begin{subfigure}[b]{0.19\textwidth}
    \includegraphics[width=\textwidth]{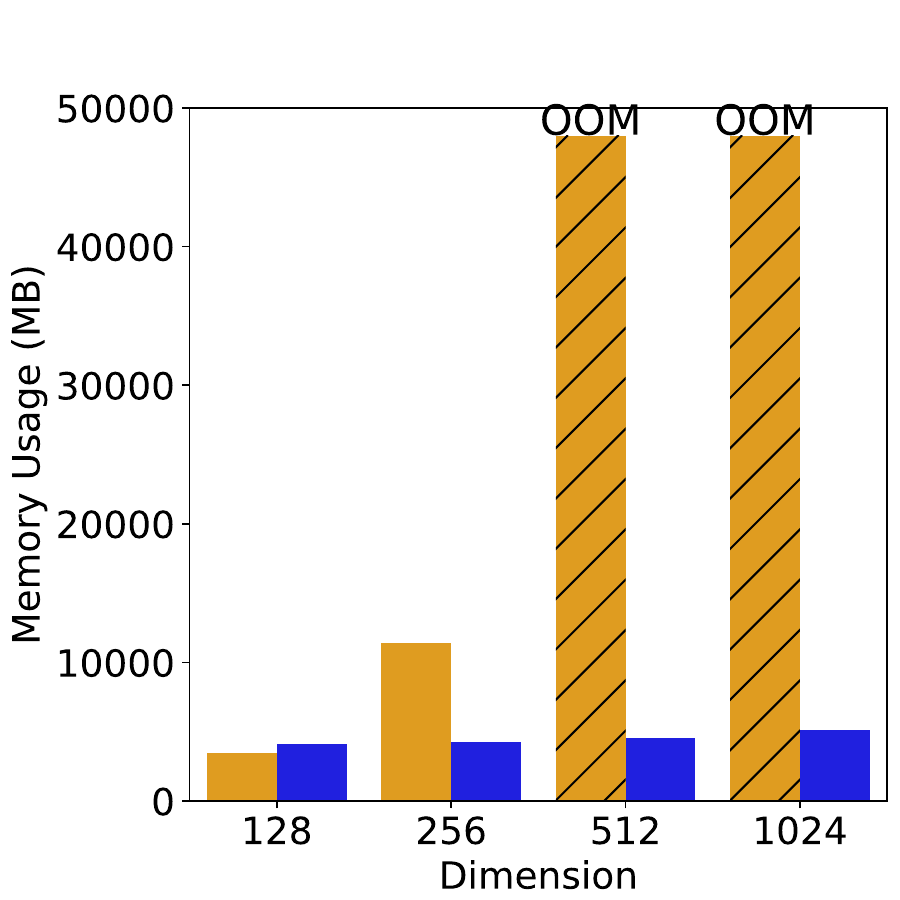}
    \caption{Mushroom}
  \end{subfigure}
  \begin{subfigure}[b]{0.19\textwidth}
    \includegraphics[width=\textwidth]{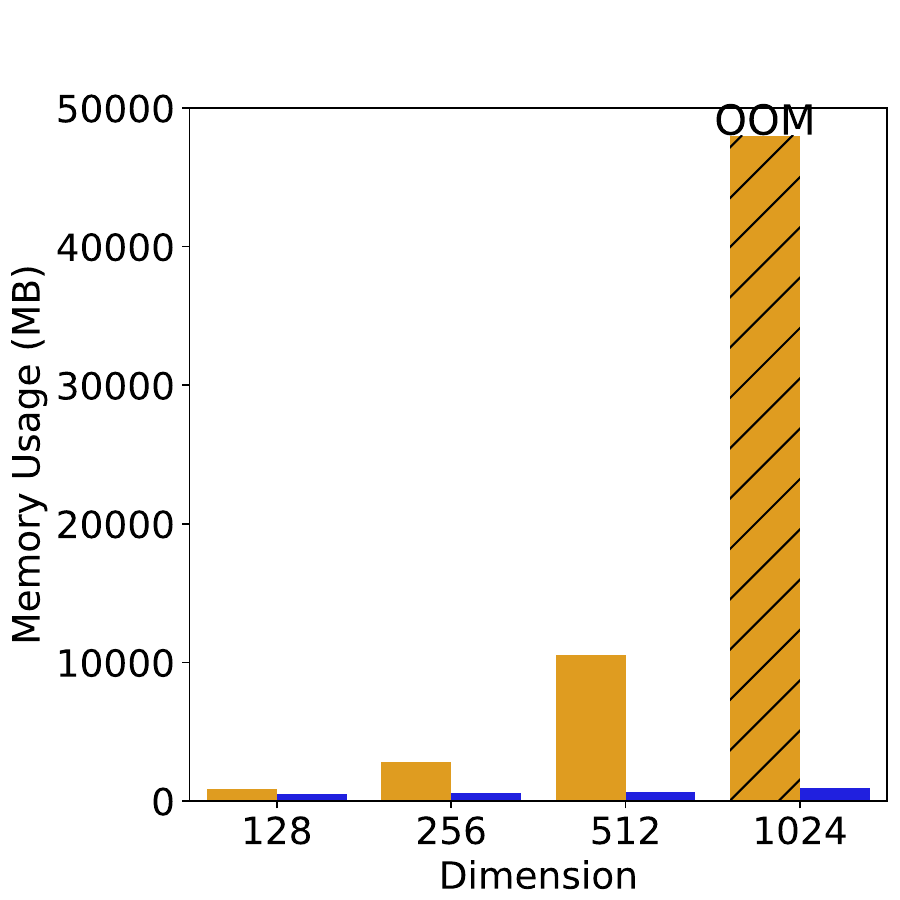}
    \caption{NTU2012}
  \end{subfigure}
  \begin{subfigure}[b]{0.19\textwidth}
    \includegraphics[width=\textwidth]{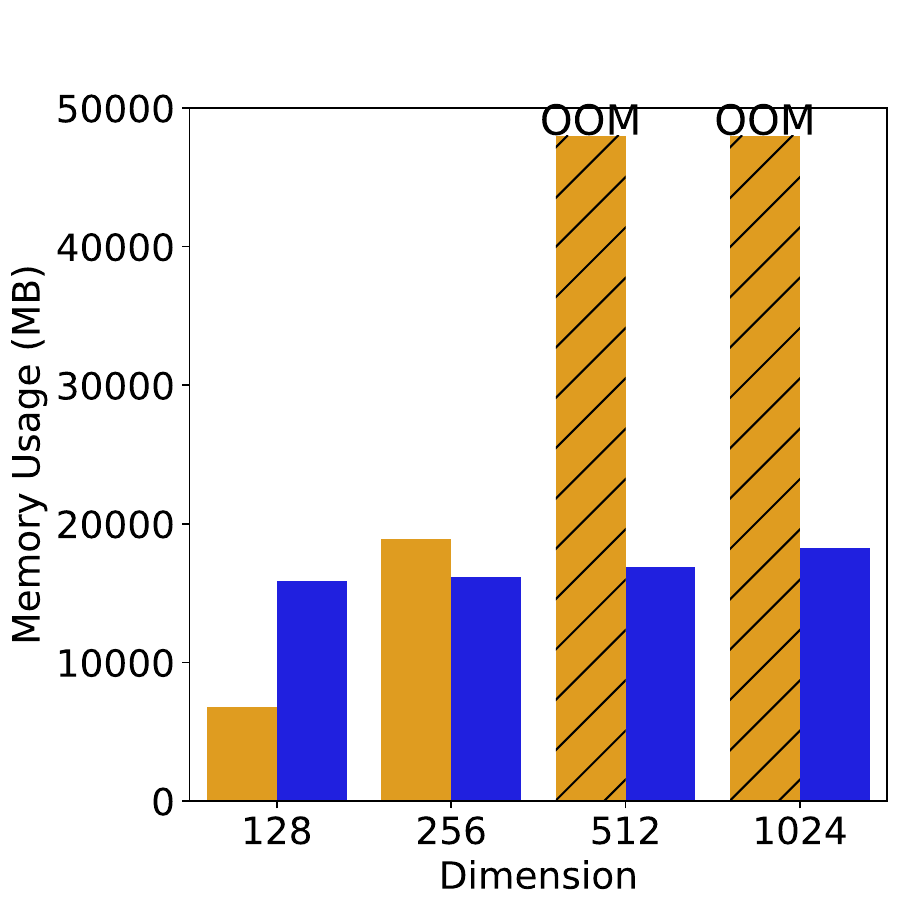}
    \caption{ModelNet40}
  \end{subfigure}
  \caption{Illustration of the GPU memory usage comparison between HyFi and TriCL, focusing on how they differ in memory consumption in various dimensions, where dimension denotes the encoder output dimension and the project head output dimension, a bar with a diagonal pattern means `Out of Memory'.}
  \label{fig:memory}
\end{figure*}

\begin{table*}[t!]
\centering
\caption{Ablation study exploring various configurations of the contrastive learning framework `w/o' stands for `without'.}
\resizebox{\textwidth}{!}{
\renewcommand{\arraystretch}{1.5}
\begin{tabular}{lccccccccccc}
\hline
 & {Cora-C} & {Citeseer} & {Pubmed} & {Cora-A} & {DBLP} & {Zoo} & {20News} & {Mushroom} & {NTU2012} & {ModelNet40} & {A.R.$\downarrow$} \\ \hline\hline
\textbf{HyFi} & \textbf{81.48\textpm1.5} & \underline{72.57\textpm1.1} & \textbf{83.82\textpm0.6} & \textbf{79.33\textpm1.2} & \underline{90.41\textpm0.2} & \textbf{80.02\textpm10.9} & \textbf{79.76\textpm0.3} & \underline{99.79\textpm0.2} & \textbf{75.10\textpm2.6} & \textbf{97.38\textpm0.1} & \textbf{1.4}\\ 
w/o weak positive pair & 70.89\textpm3.6 & 60.12\textpm3.6 & 79.91\textpm0.9 & 75.75\textpm1.1 & 88.46\textpm0.1 & 79.79\textpm10.9 & 79.69\textpm0.3 & \textbf{99.87\textpm0.1} & 74.88\textpm2.6 & 97.31\textpm0.1 & 4.2\\ 
w/o positive pair & 77.96\textpm1.7 & 71.97\textpm1.1 & 83.59\textpm0.6 & 76.65\textpm1.4 & 87.62\textpm0.6 & \underline{79.94\textpm11.0} & \underline{79.74\textpm0.3} & 99.79\textpm0.1 & 74.91\textpm2.6 & \underline{97.35\textpm0.1} & 3.2\\ 
w/o weak weight & 80.44\textpm1.3 & \textbf{72.61\textpm1.1} & \underline{83.79\textpm0.7} & 78.44\textpm1.1 & \textbf{90.54\textpm0.1} & 79.77\textpm10.9 & \textbf{79.76\textpm0.3} & 99.76\textpm0.2 & 74.52\textpm2.5 & 97.32\textpm0.1 & 2.9\\ 
w/o edge loss & \underline{81.25\textpm1.4} & 72.42\textpm1.2 & 83.77\textpm0.7 & \underline{79.04\textpm1.1} & 90.31\textpm0.2 & 79.79\textpm11.0 & \underline{79.74\textpm0.3} & \underline{99.79\textpm0.2} & \underline{74.94\textpm2.6} & 97.26\textpm0.1 & \underline{2.8}\\ 
\hline
\end{tabular}}
\label{tab:ablation}
\end{table*}

\subsubsection{Training time}

Table \ref{tab:speed} shows the results of training speed comparisons between the HyFi model and the TriCL model across 10 experimental datasets, highlighting HyFi's significant efficiency advantage. HyFi consistently exhibits faster training times compared to TriCL on these datasets. The disparity in training speeds is particularly pronounced in larger datasets such as PubMed and DBLP. HyFi completes training in just 69.22 seconds and 481.86 seconds for these datasets, respectively, significantly faster than TriCL, which takes 350.68 seconds and 1145.62 seconds. Such results underscore the efficiency of HyFi, especially in scenarios where rapid model training is essential. The consistent speed advantage across various datasets highlights the optimized nature of the HyFi model.

\subsubsection{Memory cost}

As shown in Figure \ref{fig:memory}, the HyFi model shows significant improvements in GPU memory efficiency compared to the TriCL model across the data sets, without exception. Unlike TriCL, which shows a steep increase in memory consumption at larger dimensions, HyFi maintains a relatively moderate growth in GPU memory consumption. This difference becomes more pronounced at larger dimensions, where TriCL experiences a sensitive increase in memory consumption, leading to out-of-memory problems on datasets such as DBLP, Mushroom, NTU2012 and ModelNet40. HyFi, on the other hand, never runs out of memory on any dataset.

The high efficiency of HyFi is due to two main factors: the strategy of avoiding augmentation in the hypergraph topology and the simplification of the contrastive learning process from three losses in TriCL to only two in HyFi (no need for membership-level loss). As a result, the significant difference in memory consumption between HyFi and TriCL demonstrates the strength of HyFi in handling large datasets or data with high dimensionality of node features.

\begin{table*}[t!]
\centering
\caption{Performance of different augmentation method on various datasets.}
\resizebox{\textwidth}{!}{
\renewcommand{\arraystretch}{1.5}
\begin{tabular}{ccccccccccccc}
\hline
Augmentation Type & Method & {Cora-C} & {Citeseer} & {Pubmed} & {Cora-A} & {DBLP} & {Zoo} & {20News} & {Mushroom} & {NTU2012} & {ModelNet40} & {A.R.$\downarrow$}\\ \hline \hline
\multirow{3}{*}{Add Noise} & Gaussian Noise & \textbf{81.48\textpm1.5} & \textbf{72.57\textpm1.1} & \underline{83.82\textpm0.6} & \textbf{79.33\textpm1.2} & \textbf{90.41\textpm0.2} & \textbf{80.02\textpm10.9} & \textbf{79.76\textpm0.3} & 99.79\textpm0.2 & \textbf{75.10\textpm2.6} & \textbf{97.38\textpm0.1} & \textbf{1.4}\\ 
 & Uniform Noise & \underline{81.40\textpm1.2} & \underline{72.51\textpm1.1} & \textbf{84.03\textpm0.7} & \underline{77.61\textpm1.2} & \underline{90.37\textpm0.2} & 79.65\textpm10.8 & \textbf{79.76\textpm0.3} & 99.79\textpm0.1 & 74.52\textpm2.5 & 97.32\textpm0.1 & \underline{2.9}\\ 
 & Bernoulli Noise & 76.78\textpm1.5 & 71.17\textpm1.3 & 82.54\textpm0.7 & 76.78\textpm1.2 & 90.01\textpm0.2 & 79.60\textpm10.8 & \textbf{79.76\textpm0.3} & 99.78\textpm0.1 & 74.40\textpm2.5 & 97.29\textpm0.1 & 5.2\\ \hline
\multirow{3}{*}{Drop}  & Incidence & 78.81\textpm1.5 & 72.22\textpm1.2 & 83.59\textpm0.6 & 77.54\textpm1.3 & 89.12\textpm0.4 & 79.91\textpm10.9 & 79.73\textpm0.3 & \underline{99.82\textpm0.1} & 75.05\textpm2.6 & \underline{97.37\textpm0.1} & 3.5\\ 
 & Node & 78.89\textpm1.5 & 72.25\textpm1.2 & 83.58\textpm0.6 & 77.57\textpm1.3 & 89.05\textpm0.4 & 79.90\textpm10.9 & 79.73\textpm0.3 & \underline{99.82\textpm0.1} & \underline{75.08\textpm2.6} & \textbf{97.38\textpm0.1} & 3.1\\ 
 & Hyperedge & 78.71\textpm1.5 & 72.25\textpm1.1 & 83.58\textpm0.6 & 77.30\textpm1.4 & 88.86\textpm0.4 & \underline{79.98\textpm11.0} & 79.73\textpm0.3 & \textbf{99.83\textpm0.1} & 75.04\textpm2.6 & \underline{97.37\textpm0.1} & 3.7\\ \hline
\end{tabular}}
\label{tab:augmentation_results}
\end{table*}

\begin{figure*}[thbp!]
  \centering
  \begin{subfigure}[b]{0.19\textwidth}
    \includegraphics[width=\textwidth]{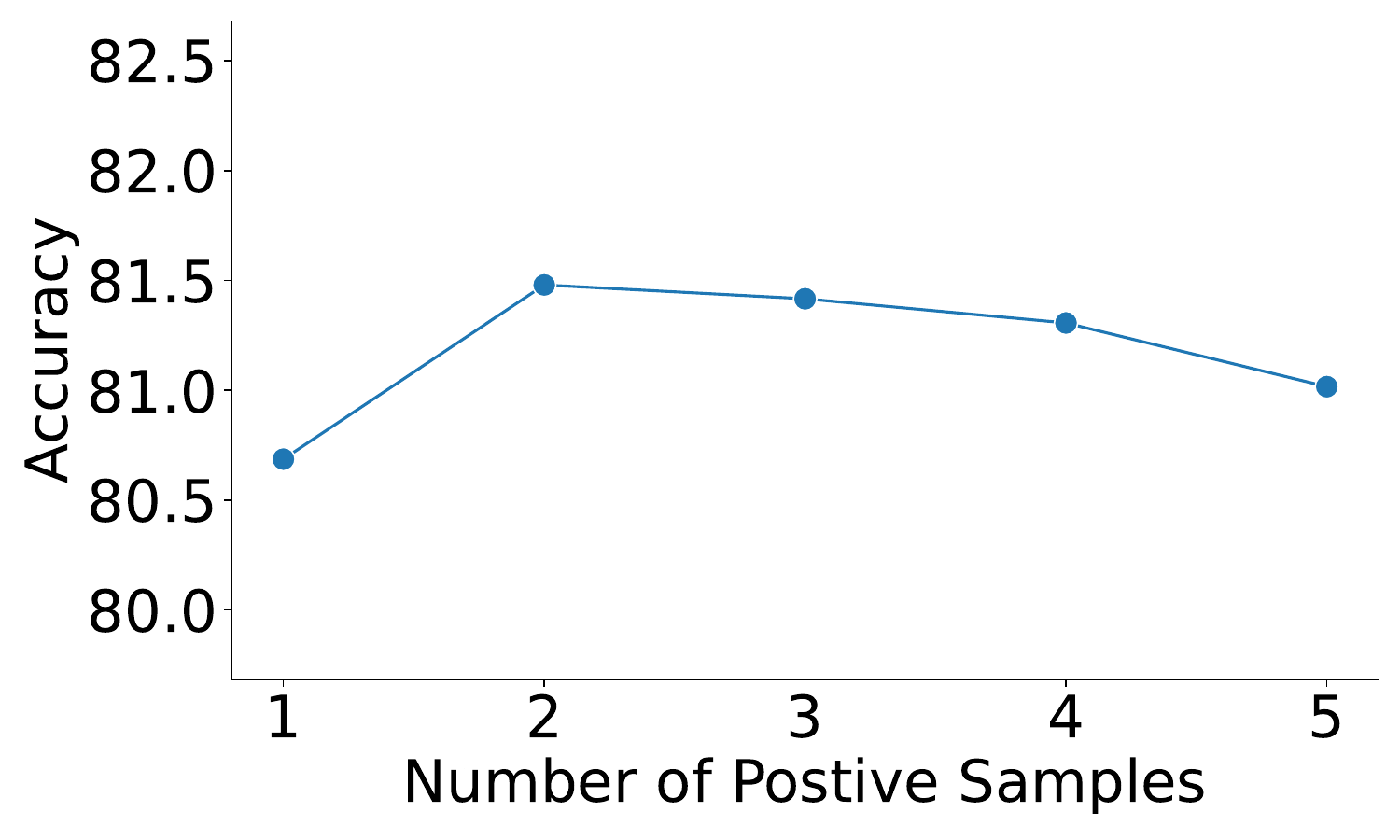}
    \caption{Cora-C}
  \end{subfigure}
  \begin{subfigure}[b]{0.19\textwidth}
    \includegraphics[width=\textwidth]{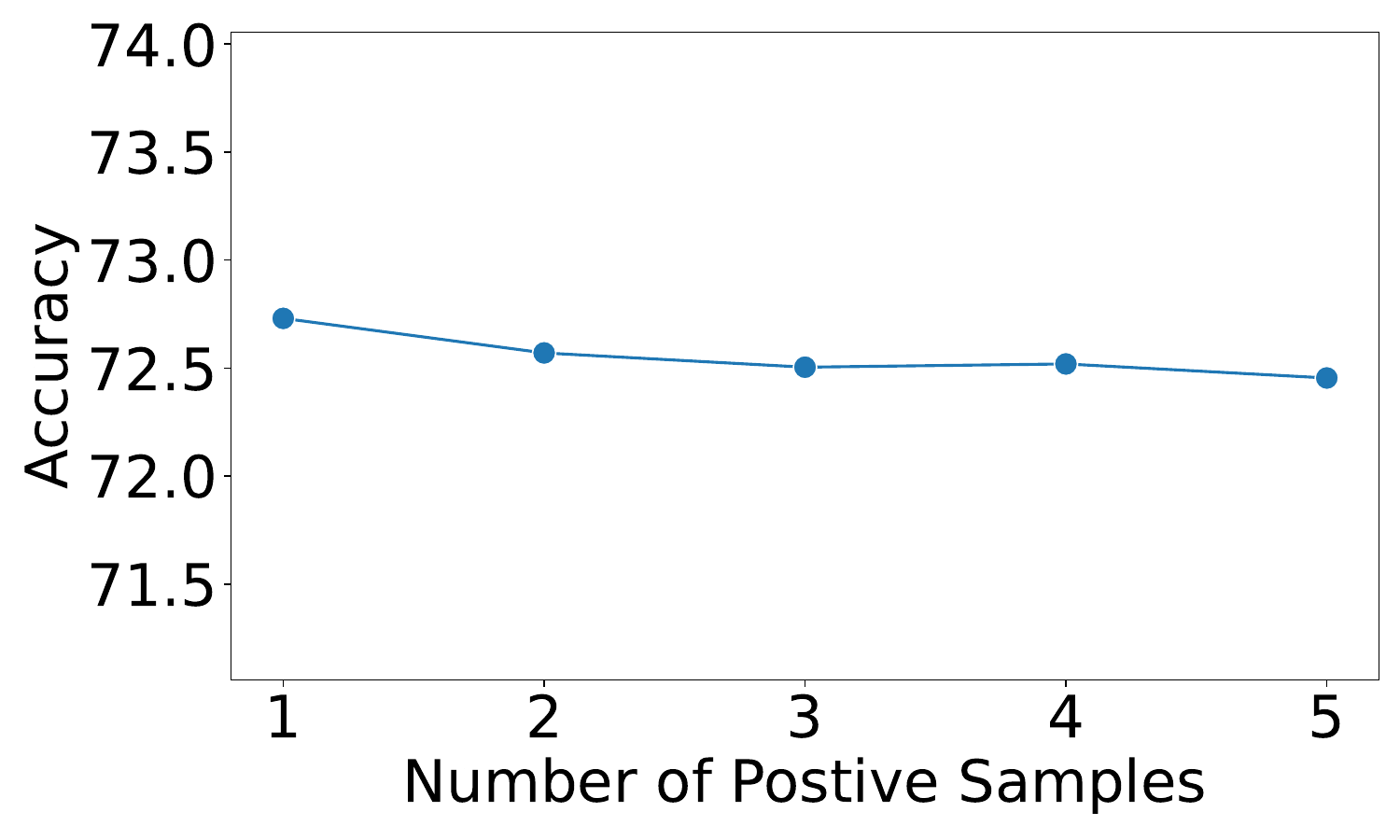}
    \caption{Citeseer}
  \end{subfigure}
  \begin{subfigure}[b]{0.19\textwidth}
    \includegraphics[width=\textwidth]{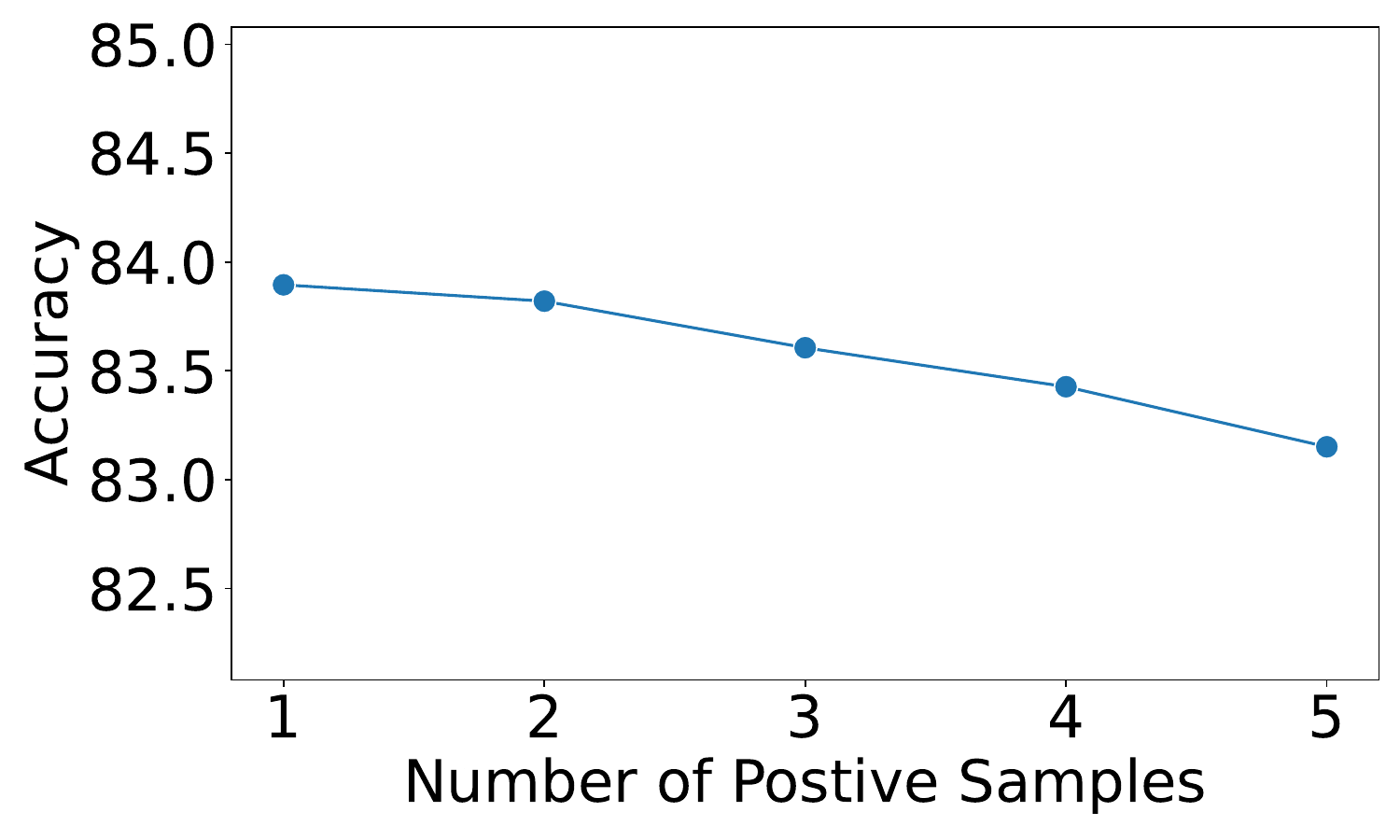}
    \caption{Pubmed}
  \end{subfigure}
  \begin{subfigure}[b]{0.19\textwidth}
    \includegraphics[width=\textwidth]{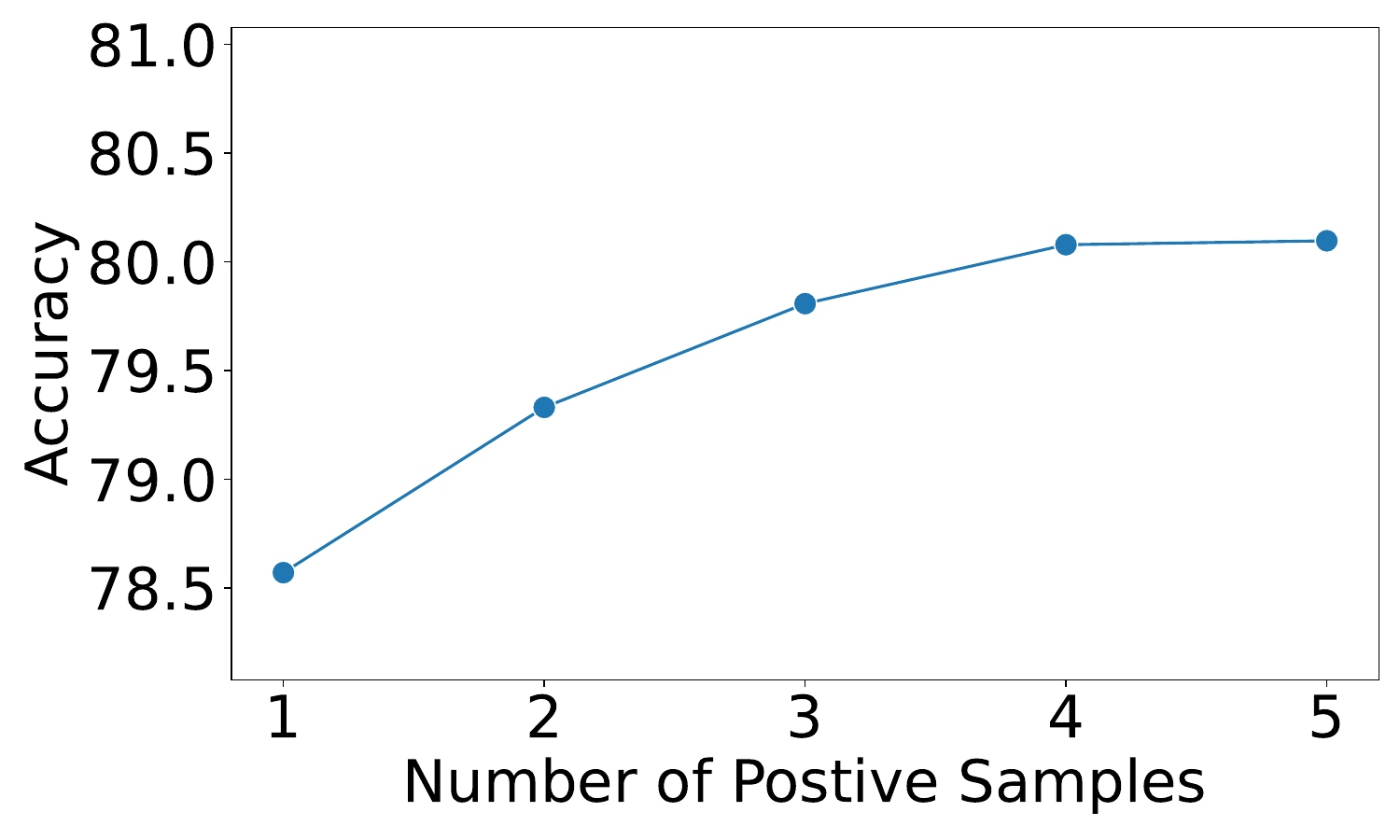}
    \caption{Cora-A}
  \end{subfigure}
  \begin{subfigure}[b]{0.19\textwidth}
    \includegraphics[width=\textwidth]{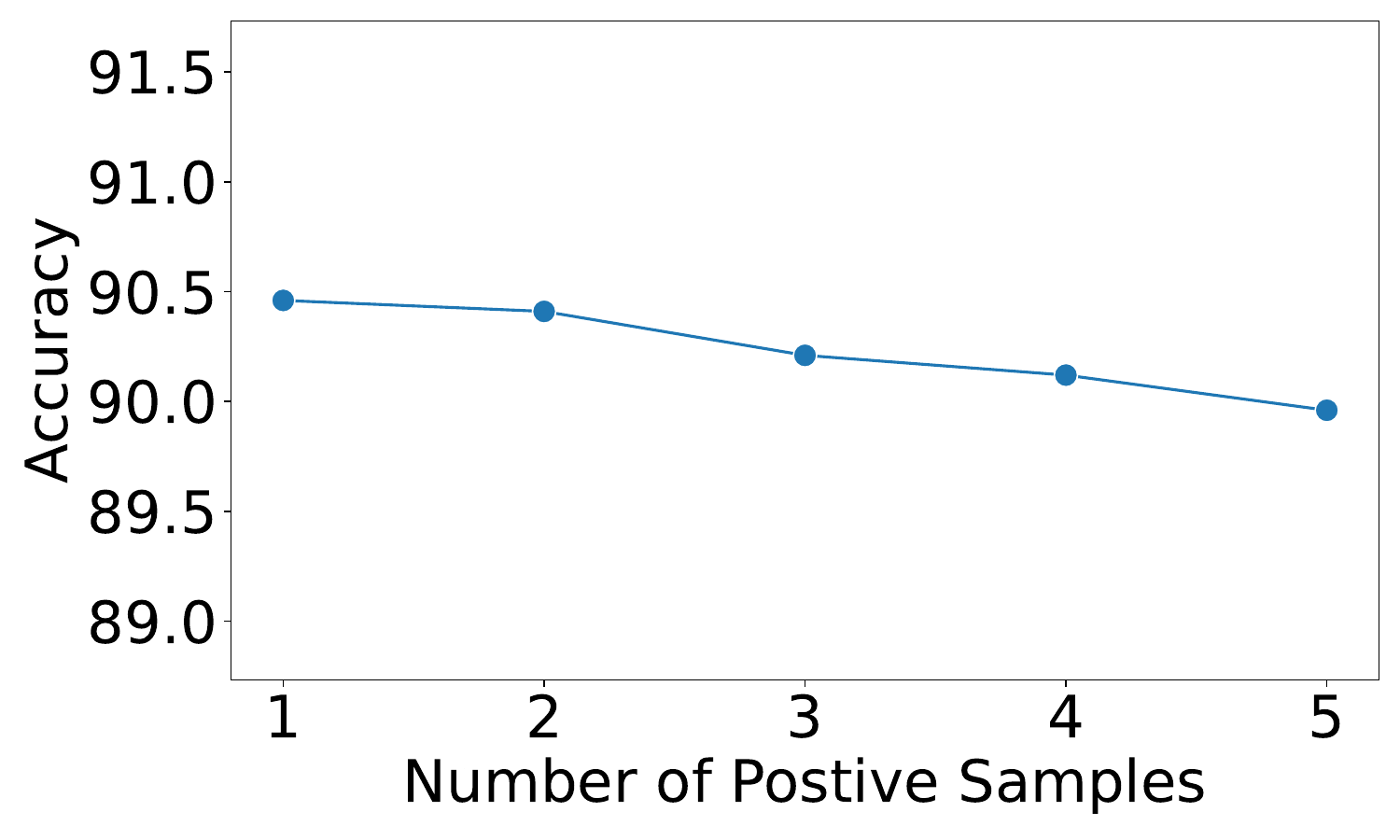}
    \caption{DBLP}
  \end{subfigure}
  \begin{subfigure}[b]{0.19\textwidth}
    \includegraphics[width=\textwidth]{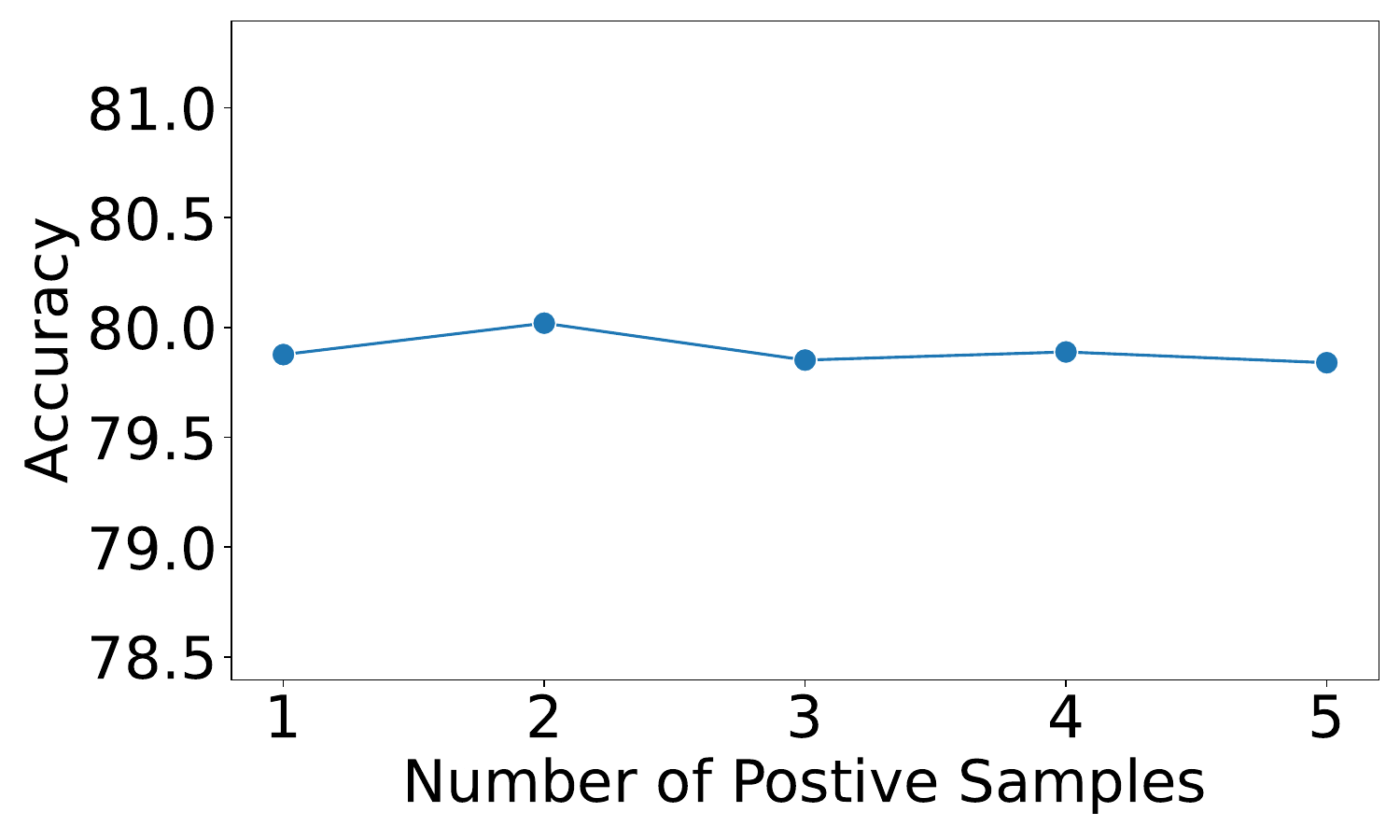}
    \caption{Zoo}
  \end{subfigure}
  \begin{subfigure}[b]{0.19\textwidth}
    \includegraphics[width=\textwidth]{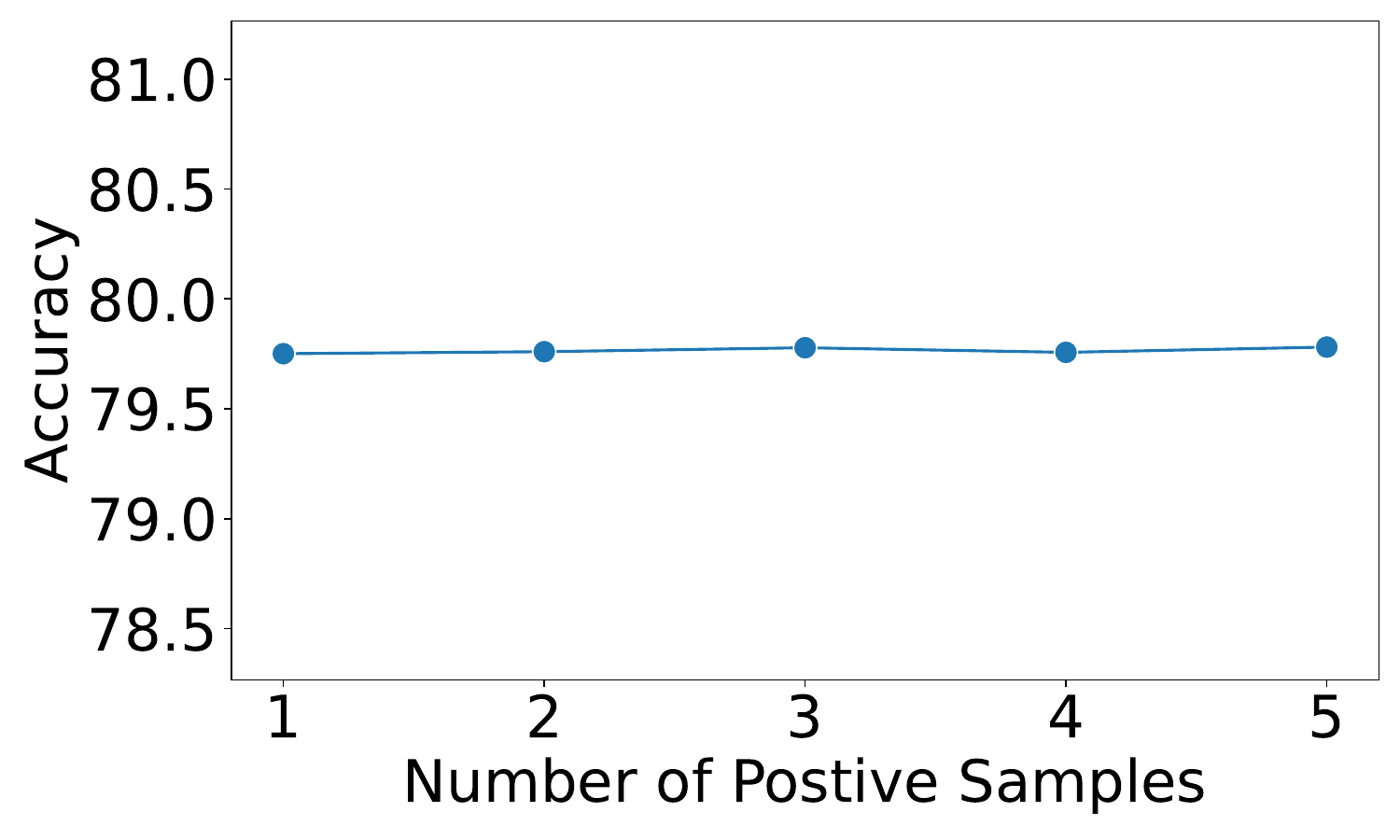}
    \caption{20News}
  \end{subfigure}
  \begin{subfigure}[b]{0.19\textwidth}
    \includegraphics[width=\textwidth]{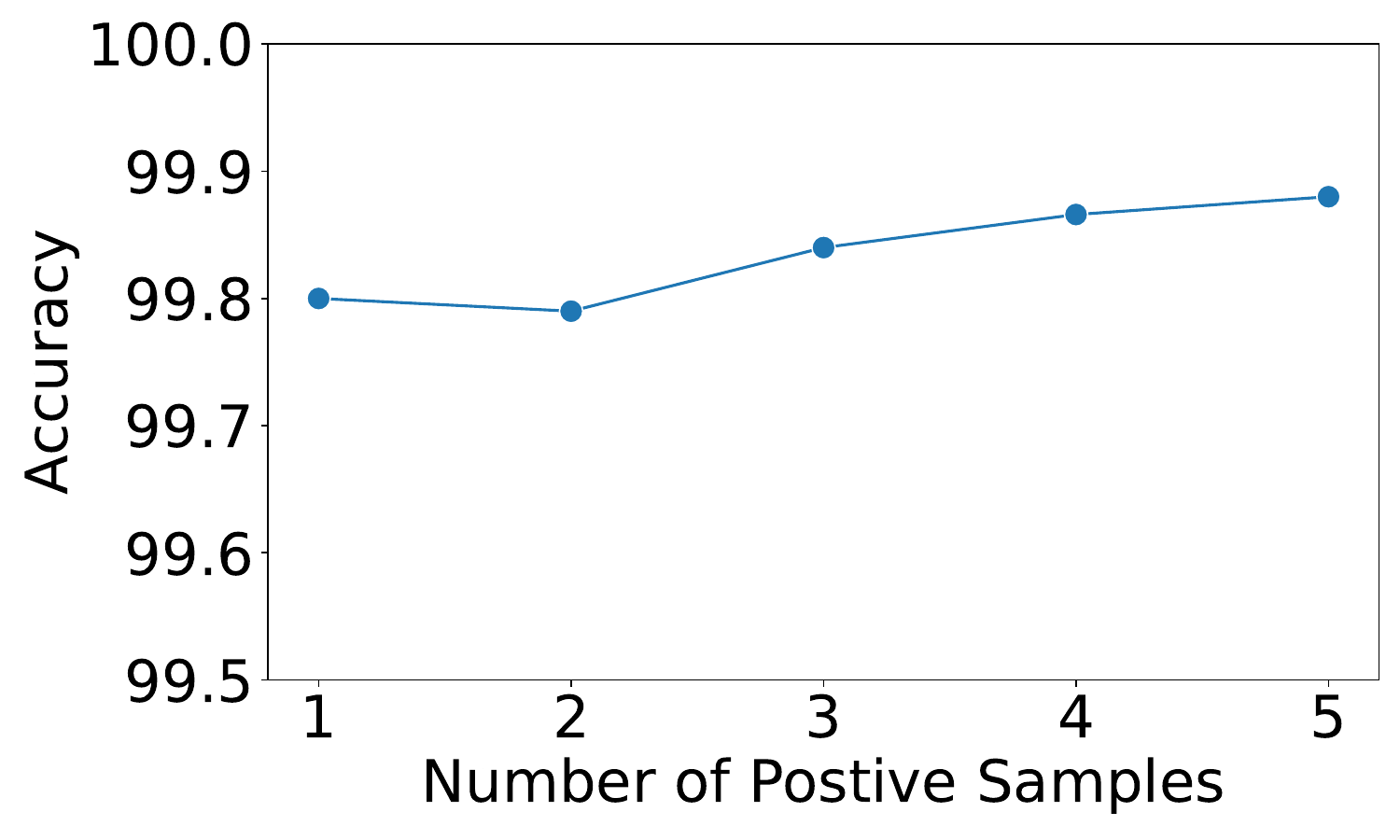}
    \caption{Mushroom}
  \end{subfigure}
  \begin{subfigure}[b]{0.19\textwidth}
    \includegraphics[width=\textwidth]{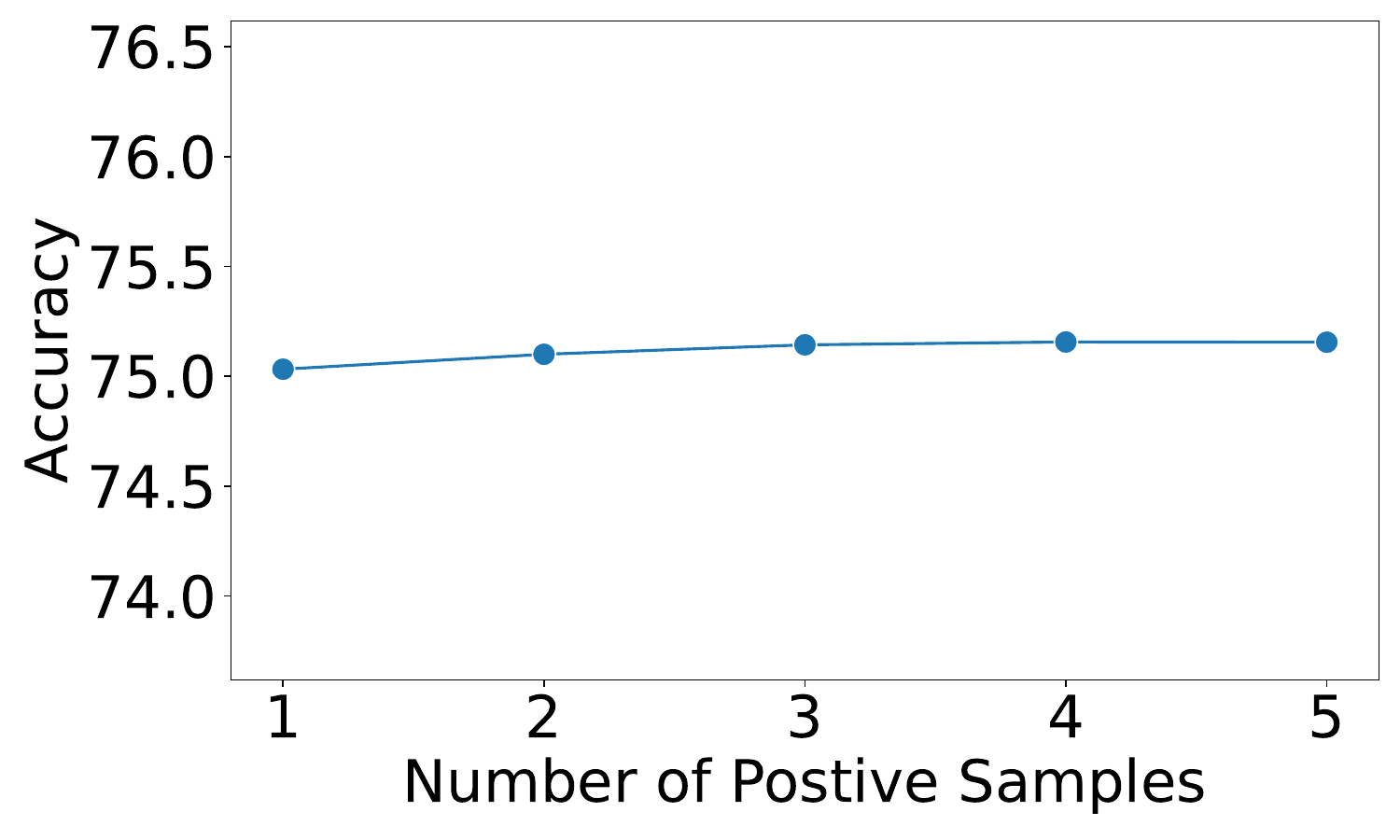}
    \caption{NTU2012}
  \end{subfigure}
  \begin{subfigure}[b]{0.19\textwidth}
    \includegraphics[width=\textwidth]{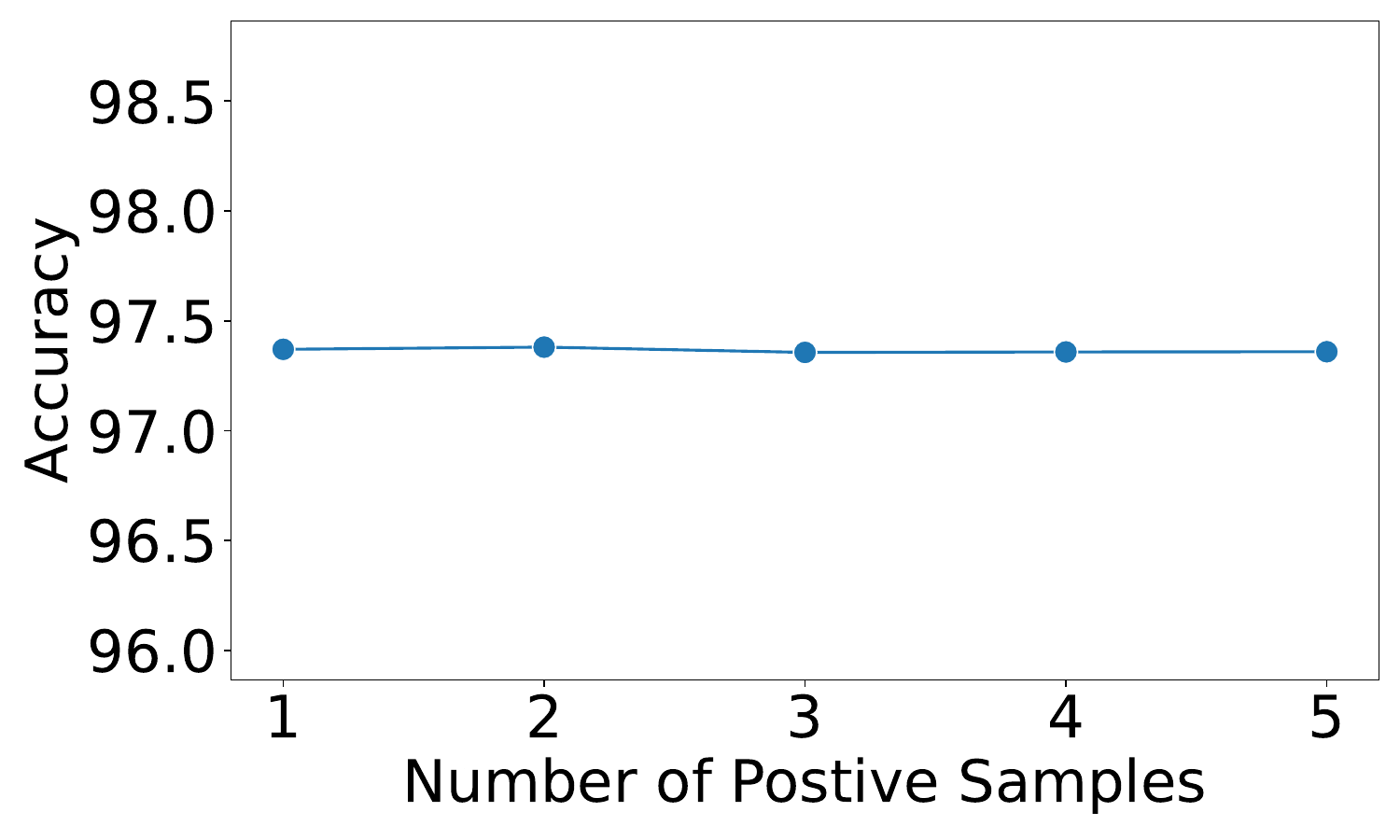}
    \caption{ModelNet40}
  \end{subfigure}
  \caption{Illustration of the performance of node classification as the number of positive samples. The number of noise views is equal to the number of positive samples in the contrastive learning.}
  \label{fig:num_pos}
\end{figure*}

\subsection{Ablation Study (RQ3)} 

The ablation study, detailed in Table \ref{tab:ablation}, evaluates the efficacy of various components on the performance of the HyFi. The study considers four different configurations: a model that does not use weak positive pairs (only positive and negative pairs), a model that does not use positive pairs (only weak positive and negative pairs), a model that does not utilize weights for weak positive pairs, and a model with no edge-level loss.

The results for the model that does not use weak positive pairs show a significant decrease in performance, which is more pronounced than the decrease observed in the model without positive pairs. Except for a few datasets, the performance of the model using only weak positive pairs is almost as good as when using both positive and weak positive pairs. These results demonstrate that the commonality of sharing the same group can be very important in hypergraph contrastive learning. We also observe a slight performance drop when weights for weak positive pairs are not used or when edge-level loss is not utilized. The results indicate that weighted contributions further enhance the model's efficiency. Additionally, we observe that edge-level loss also helps to improve the model's performance.

\subsection{Types of Augmentation (RQ4)}




We experimented with different augmentation methods to determine the suitability of adding Gaussian noise to node features to generate a positive sample. We experimented with various types of noise (Gaussian, Uniform, Bernoulli). Bernoulli noise perturbs the node feature at values of 0 and 1, which perturbs the node feature significantly compared to Gaussian noise and uniform noise. We also compared our method with common graph augmentation techniques, such as drop augmentation, which involve removing the incidence, node, or hyperedge, thereby modifying the topology of the hypergraph. We used a drop rate of 20\%.

Among all graph augmentation methods, adding Gaussian noise achieved the best average ranking. However, since graph augmentation in HyFi is basically aimed at creating positive pairs, adding some noise to the node features is enough to create positive pairs. However, methods that modify the topology of the graph performed poorly in terms of average ranking, suggesting that graph augmentation methods that change the topology are less effective at generating positive samples. Furthermore, the topology of the graph relies heavily on randomness, suggesting the possibility of losing important information. For example, the dropping of a community called family will result in a significant loss of information for some nodes. As such, HyFi's graph augmentation method has the advantage of being able to be used on a variety of graphs without worrying about information loss due to variations in the topology of the graph. Therefore, HyFi, which does not disturb the topology of the graph, can be utilized on a variety of graphs and in a variety of downstream tasks.

\subsection{Effect of Positive sample (RQ5)}

Figure \ref{fig:num_pos} shows how performance varies with the number of positive samples. In HyFi, we use 2 positive samples: the number of positive samples is equal to the number of noise views, and the number of positive samples is considered a hyperparameter. We found that the number of positive samples has little effect on the overall performance. We observed less than a 1\% difference in performance for most datasets, and it was difficult to find a correlation between the number of positive samples and performance. This indicates that HyFi's performance is not driven by the relationship of weak positive pairs, and that the relationship of positive pairs is not critical to HyFi's performance.

\section{Conclusion}
\label{sec:conclusions}
This study has successfully demonstrated the efficacy of the HyFi model in hypergraph-based contrastive learning, demonstrating its ability to capture the nuanced relationships inherent in hypergraph results in higher quality embeddings, and superior performance in node classification task across a variety of datasets. Our novel augmentation technique developed for hypergraphs preserves the original topology while adding noise to node features, ensuring the integrity of the hypergraph structure. Furthermore, the contrastive learning method introduced in this study effectively utilizes the homophily characteristic of hypergraphs, providing a more effective approach than existing HGCL methods. Also, the proposed HyFi model stands out for its efficiency in processing speed and memory usage, making it an ideal solution for various applications. 

\section{Acknowledgement}
This work was supported by the Technology Innovation Program funded By the Ministry of Trade, Industry \& Energy (MOTIE) (No.20022899) 

\bibliographystyle{IEEEtranS}

\bibliography{references}

\end{document}